\documentclass{article}
\usepackage{arxiv}
\usepackage{url}

\usepackage[colorlinks,bookmarksopen,bookmarksnumbered,citecolor=magenta,urlcolor=magenta]{hyperref}

\usepackage{natbib}
\usepackage{soul}
\usepackage{amsmath}
\usepackage{amssymb}
\usepackage{tikz}
\usepackage{multirow}
\usepackage{adjustbox}

\usepackage{hyperref}

\usepackage{etoolbox}
\usetikzlibrary{arrows,decorations.markings}

\usepackage[singlelinecheck=off,justification=raggedright]{subcaption}

\newcommand{\toprule}{\hline}
\newcommand{\midrule}{\hline}
\newcommand{\bottomrule}{\hline}
\newcommand{\trust}{\ensuremath{\tau^a}}

\newcommand{\robot}{\ensuremath{a}}
\newcommand{\taskfeat}{\ensuremath{\mathbf{x}}}
\newcommand{\taskz}{\ensuremath{\mathbf{z}}}
\newcommand{\performance}{\ensuremath{c^\robot}}
\newcommand{\perfz}{\ensuremath{\mathbf{c}^\robot}}
\newcommand{\btheta}{\ensuremath{\boldsymbol{\theta}}}

\newcommand{\mat}[1]{\mathbf{#1}}

\newcommand\BibTeX{{\rmfamily B\kern-.05em \textsc{i\kern-.025em b}\kern-.08em
T\kern-.1667em\lower.7ex\hbox{E}\kern-.125emX}}

\begin{document}

\title{Multi-Task Trust Transfer for\\Human-Robot Interaction}

\author{Harold Soh\thanks{Email: \url{harold@comp.nus.edu.sg}. Published in \textit{The International Journal of Robotics Research} (IJRR), \url{https://doi.org/10.1177/0278364919866905}. An earlier conference version appeared in Robotics: Science and Systems (RSS) 2018, \url{http://www.roboticsproceedings.org/rss14/p33.html}}, Yaqi Xie, Min Chen, \textnormal{and} David Hsu\\Department of Computer Science\\School of Computing\\National University of Singapore}

\maketitle

\begin{abstract}
Trust is essential in shaping human interactions with one another and with robots. This paper discusses how human trust in robot capabilities transfers across multiple tasks. We first present a human-subject study of two distinct task domains: a Fetch robot performing household tasks and a virtual reality simulation of an autonomous vehicle performing driving and parking maneuvers. The findings expand our understanding of trust and inspire new differentiable models of trust evolution and transfer via latent task representations: (i) a rational Bayes model, (ii) a data-driven neural network model, and (iii) a hybrid model that combines the two. Experiments show that the proposed models outperform prevailing models when predicting trust over unseen tasks and users. These results suggest that (i) task-dependent functional trust models capture human trust in robot capabilities more accurately, and (ii) trust transfer across tasks can be inferred to a good degree. The latter enables trust-mediated robot decision-making for fluent human-robot interaction in multi-task settings.
\end{abstract}

\section{Introduction}
\label{sec:intro}

As robots enter our homes and workplaces, interactions between humans and
robots become ubiquitous. \textit{Trust} plays a prominent role in shaping these interactions and
directly affects the degree of autonomy rendered to
robots~\citep{sheridan1984research}. This has led to significant
efforts in conceptualizing and measuring human trust in robots
and automation~\citep{muir1994trust,lee1994trust,Castelfranchi2010,yang2017evaluating}. 

A crucial gap, however,  remains in understanding 
when and how human trust in robots \textit{transfers} across multiple tasks based on the {human's prior knowledge of robot task capabilities and past experiences}. Understanding trust in the multi-task
setting is crucial as robots transition from single-purpose machines in
controlled environments---such as factory floors---to general-purpose partners
performing diverse functions.
{The mathematical formalization of trust across tasks lays the foundation of trust-mediated robot decision making for fluent human-robot
interaction}. In particular, it leads to robot policies that mitigate under-trust or over-trust by humans when interacting with robots~\citep{Chen2018Hri}.

In this work, we take a first step towards formalizing trust transfer across tasks for human-robot interaction. 
We adopt the definition of trust as a {psychological
  attitude}~\citep{Castelfranchi2010} and {focus on \emph{trust in robot
  capabilities}, i.e., the belief in a robot's competence to complete a task. Capability is a primary factor in determining overall trust in
robots}~\citep{muir1994trust}, { and this work investigates how trust in robot capabilities varies and transfers across a range of tasks.} 

\begin{figure}
	\centering
	\includegraphics[width=0.7\textwidth]{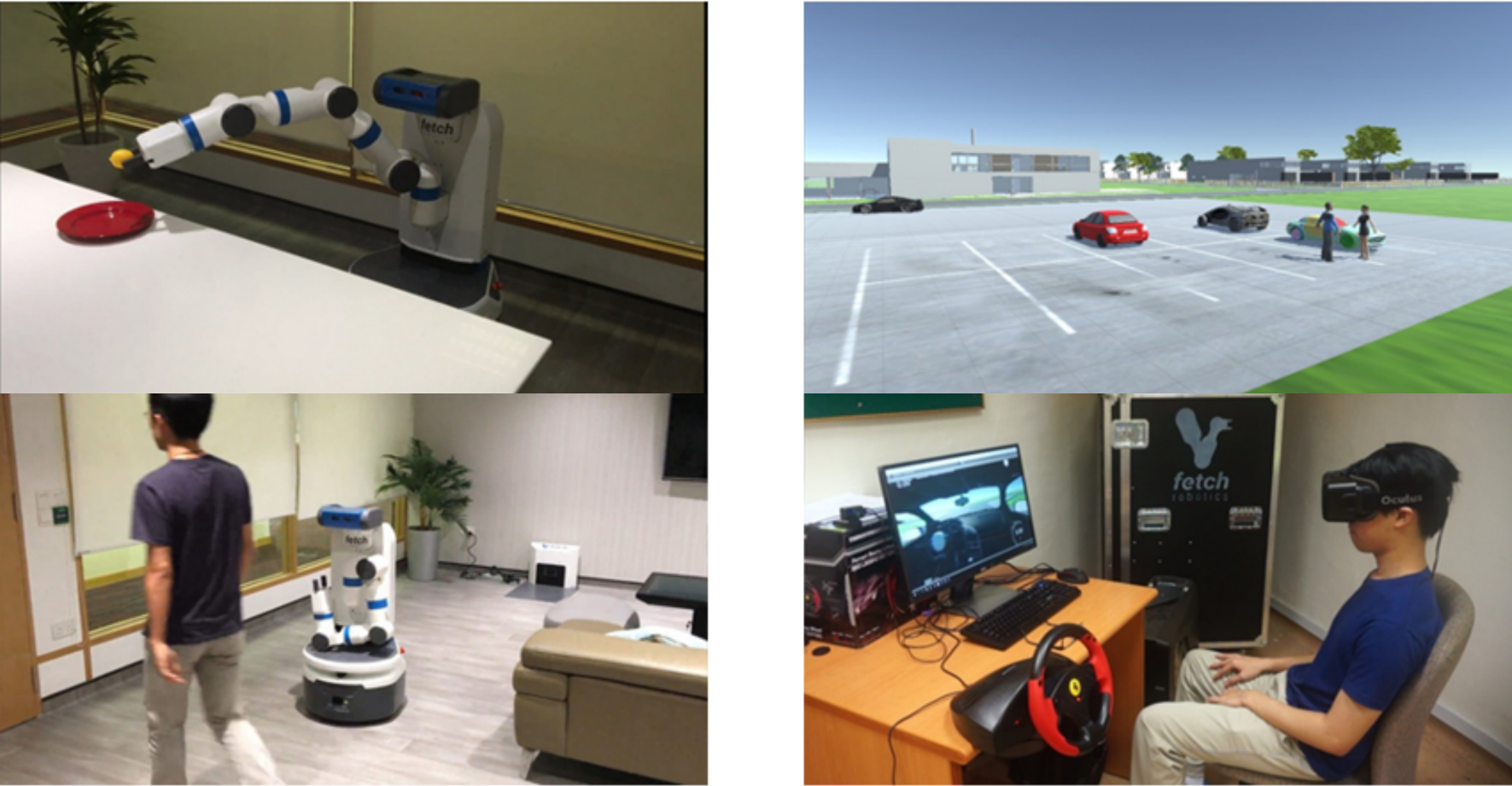}
	\caption{Experiment Task Domains: (Left) Household tasks with the Fetch Research Robot picking and placing objects (top left) and indoor navigation (bottom left) (Right) Autonomous Driving tasks in the Virtual Reality simulation system. Tasks included parking and various navigation scenarios.
	}
	\label{fig:domains}
\end{figure}

{Our first contribution is a human-subject study ($n=32$) where our goal is to uncover the role of task similarity and difficulty in the formation and dynamics of trust}. We present results in two  task domains: household tasks and autonomous driving (Fig. \ref{fig:domains}). The two disparate domains allow us to validate the robustness  of our findings. We show that inter-task trust transfer depends on perceived task similarity, difficulty, and observed robot performance. These results are consistent across both domains, even though the robots and the contexts are markedly different: the household domain, involves a Fetch robot that navigates and that picks and places everyday objects, while the driving domain involves a virtual reality (VR) simulation of an autonomous vehicle performing driving and parking maneuvers. To our knowledge, this is the first work showing concrete evidence for trust transfer across tasks in the context of human-robot interaction. {We have made our data and code freely available online for further research}~\citep{SuppMat}.

Based on our experimental findings, we propose to conceptualize trust as a \emph{context-dependent latent dynamic function}. This viewpoint captures the main findings of our experiments and is supported by prior socio-cognitive research showing the dependence of trust  on task properties and on the agent to be trusted~\citep{Castelfranchi2010}. We focus on  characterizing the \emph{structure} of this ``trust function'' and its \emph{dynamics}, i.e., how it changes with observations of robot performance across tasks. An earlier version of this paper~\citep{SohRSS18} presents two formal models: (i) a Bayesian Gaussian process (GP)~\citep{rasmussen06} model, and (ii) a connectionist recurrent neural model based on recent advances in deep learning. The GP model explicitly encodes a specific assumption about how human trust evolves, via Bayes rule. In comparison, the neural model is data-driven and  places few constraints on how trust evolves with observations. Both models leverage latent task space representations (a \emph{psychological task space}) learned using word vector descriptions of tasks, e.g., ``Pick and place a glass cup''. Experiments show both models accurately predict trust across unseen tasks and users. {This paper introduces  a third model that combines the Bayesian and neural approaches and show that the hybrid model achieves improved predictions. All three models are differentiable and can be trained using standard off-the-shelf optimizers}. 

In comparison with prevailing computational models \cite[e.g.,][]{lee1994trust, xu2015optimo}, a key benefit of these trust models is their abilities to leverage inter-task structure in multi-task application settings. 
As \emph{predictive} models, they can be operationalized in decision-theoretic frameworks to calibrate trust during collaboration with human teammates~\citep{Chen2018Hri,Wang2016,Nikolaidis:2017:GMH:2909824.3020253,Huang2018}. Trust calibration is crucial for preventing over-trust that engenders unwarranted reliance in robots~\citep{Robinette2016,Singh1993}, and under-trust that can cause poor utilization~\citep{lee2004trust}.  To summarize, this paper makes the following key contributions:
\begin{itemize}
	\item A novel formalization of trust as a latent dynamic function and efficient computational differentiable models that capture and predict human trust in robot capabilities across multiple tasks; 
	\item Empirical findings from a human subjects study showing the influence of three factors on human trust transfer across tasks, i.e., perceived task similarity, difficulty, and robot performance;
	\item Systematic evaluation showing the proposed methods outperform existing methods, indicating the importance of modeling trust formation and transfer across tasks.
\end{itemize}

\section{Background and Related work}
\label{sec:relatedworks}

\begin{figure*}
	\centering
	\includegraphics[width=0.80\textwidth]{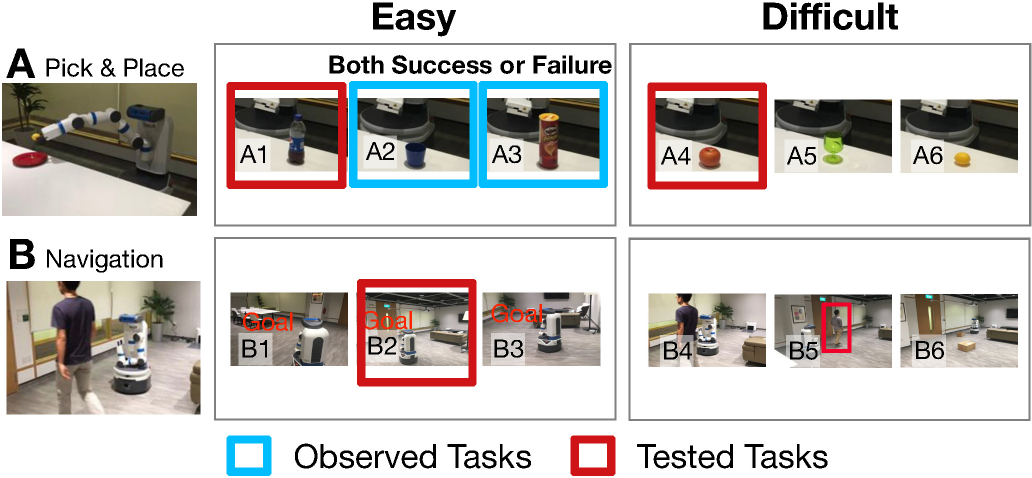}
	\caption{Trust Transfer Experiment Design. Two categories of tasks were used: (A) picking and placing different objects, and (B) navigation in a room, potentially with people and obstacles. Participants were surveyed on their trust in the robot's ability to successfully perform three different tasks (red boxes) \emph{before} and \emph{after} being shown demonstrations of two tasks. The two {demonstrated/observed} tasks were always selected from the same cell (blue boxes; cell randomly assigned, with either both successes or both failures). The tested tasks were randomly selected from three different cells---the (i) same category and difficulty level, (ii) same category but different difficulty level, and (iii) different category but same difficulty level--- compared to the observed tasks.}
	\label{fig:ExpDesign}
\end{figure*}

Research into trust in robots (and automation) is a large interdisciplinary endeavor spanning multiple fields including human-factors, psychology, and human-robot interaction. {This paper extends a prior conference version}~\citep{SohRSS18} {with additional discussion and analyses of the human-subject experiment. In addition, we include a new computational trust model --- that hybridizes the neural and Bayesian methods --- with updated experimental results and expanded discussion on the computational models, learned task-spaces, and word-based task descriptions.} In this section, we provide relevant background on trust and computational trust models.

\paragraph{Key Concepts and Definitions.} Trust is a multidimensional concept, with many factors characterizing human trust in robots, e.g., the human's technical expertise and the complexity of the robot~\citep{lee1994trust,muir1994trust,hancock2011meta}. Of these factors, two of the most prominent are the performance and integrity of the machine. Similar to existing work in robotics~\citep{Xu2016,xu2015optimo}, we assume that robots are not intentionally deceptive and focus on characterizing trust based on robot performance. We view trust as a belief in the competence and reliability of another agent to complete a task.

 There are two types of trust that differ in their situation specificity. The first is dispositional trust or trust propensity, which is an individual difference for how willing one is to trust another. The second is situational or \emph{learned} trust, that results from interaction between the agents concerned. For example, the more you use your new autonomous vehicle, the more you may learn to trust it. In this paper, we will be concerned mostly with situational trust in robots. 

\paragraph{Trust Measurement.} Trust is a latent dynamic entity, which presents challenges for measurement~\citep{Billings2012}. Previous work has derived survey instruments and methods for quantifying trust,  including binary measures~\citep{binary}, continuous measures~\citep{desai2012a,Xu2016,LEE1992}, ordinal scales~ \citep{Muir1989,Hoffman2013a,Jian2000} and an Area Under Trust Curve (AUTC) measure~\citep{Desai13,yang2017evaluating} which captures participant's trust through the entire interaction with the robot by integrating binary trust measures over time. In this paper, we use a self-reported measure of trust \citep[similar to][]{xu2015optimo} and Muir's questionnaire~\citep{muir1994trust}. 

\paragraph{Computational Models of Trust.} Previous work has explored explanatory models \citep[e.g.,][]{Castelfranchi2010,lee2004trust} and predictive models of trust. Recent models have focused on dynamic modeling, for example, a recent predictive model---OPTIMo~\citep{xu2015optimo}---is a Dynamic Bayesian Network with linear Gaussian trust updates trained on data gathered from an observational study. OPTIMo was shown to outperform an Auto-Regressive and Moving Average Value (ARMAV) model~\citep{lee1994trust}, and stepwise regression~\citep{Xu2016}. Because trust is treated as ``global'' real-valued scalar in these models, they are appropriate when tasks are fixed (or have little variation). However, as our results will show, trust can differ substantially between tasks. As such, we develop models that capture both the dynamic property of trust and its variation across tasks. We leverage upon recurrent neural networks that have been applied to a variety of sequential learning tasks~\citep[e.g.,][]{Soh2017a} and online Gaussian processes that have been previously used in   robotics~\citep{Soh2015,Soh2013a,Soh2014a}.  

\paragraph{Application of Trust Models.} Trust emerges naturally in collaborative settings. In human-robot collabation~\citep{nikolaidis2017human,Nikolaidis:2017:GMH:2909824.3020253}, trust models can be used to enable more natural  interactions. For example, \cite{Chen2018Hri} proposed a decision-theoretic model that incorporates a predictive trust model, and showed that policies that took human trust into consideration led to better outcomes. The models presented in this work can be integrated into such  frameworks to influence robot decision-making across different tasks.

\section{Human Subjects Study}
\label{sec:HumanSubjectsStudy}

In this section, we describe our human subjects study, which was designed to evaluate if and when human trust transfers between tasks. Our general intuition was that human trust generalizes and evolves in a structured manner. We specifically hypothesized that:
\begin{itemize}
\item \textbf{H1:} Trust in the robot is more similar for tasks of the same category, compared to tasks in a different category.
\item \textbf{H2:} Observations of robot performance have a greater affect on the \emph{change in human trust} over similar tasks compared to dissimilar tasks.
\item \textbf{H3:} Trust in a robot's ability to perform a task transfers more readily to easier tasks, compared to more difficult tasks.
\item \textbf{H4:} \emph{Distrust} in the robot's ability to perform a task generalizes more readily to difficult tasks, compared to easier tasks.  
\end{itemize}

\begin{figure}
\begin{center}
\includegraphics[width=0.7\textwidth]{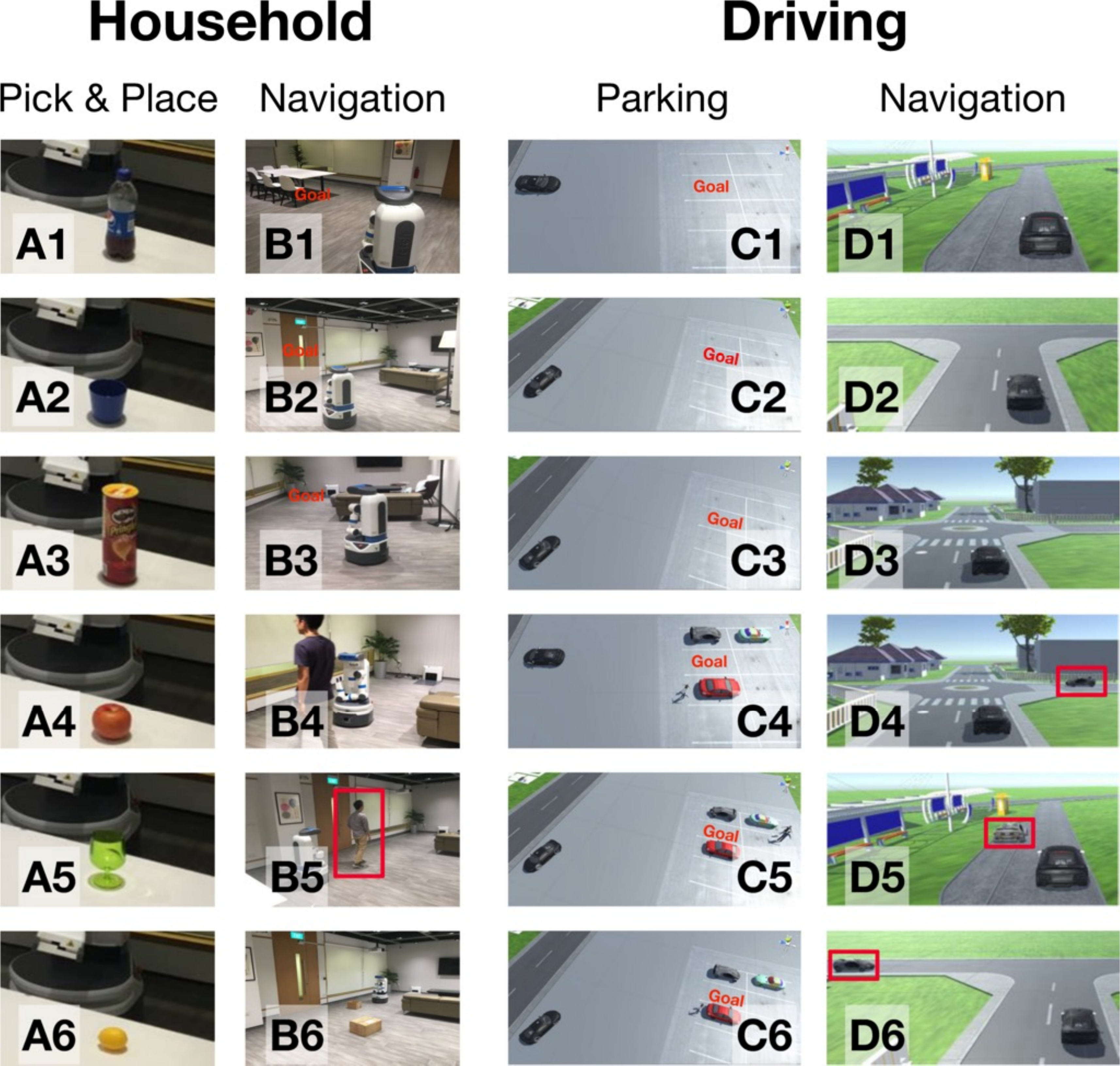}

\vspace{0.5cm}

\footnotesize
\begin{tabular}{ l| c|c | l  } 
\toprule
\textbf{Domain} & \textbf{Category} & \textbf{ID} & \textbf{Task} \\
\midrule
\multirow{12}{4.5em}{Household} & \multirow{6}{5em}{A. Pick \& Place} 
& 1 & a bottle of soda \\ 
& & 2 & a plastic cup \\ 
& & 3 & a can of chips \\
& & 4 & an apple \\ 
& & 5 & a glass cup \\ 
& & 6 & a lemon \\ 
\cline{2-4}
& \multirow{6}{5em}{B. Indoor Navigation} 
& 1 & to the table \\
& & 2 & to the door \\ 
& & 3 & to living room \\ 
& & 4 & with people moving around \\ 
& & 5 &  following a person\\ 
& & 6 & while avoiding obstacles\\ 
\cline{1-4}
\multirow{12}{4.5em}{Driving} & \multirow{6}{5em}{C. Parking} 
& 1 &  Forwards, empty lot (aligned) \\
& & 2 & Backwards, empty lot (misaligned)\\ 
& & 3 & Forwards, empty lot (misaligned)\\ 
& & 4 & Backwards, with cars (aligned) \\ 
& & 5 & Backwards, with cars (misaligned) \\ 
& & 6 & Forwards, with cars (misaligned)\\ 
\cline{2-4}
& \multirow{6}{5em}{D. Naviga-\\tion} 
& 1 & Lane merge \\
& & 2 & T-junction  \\ 
& & 3 & Roundabout  \\ 
& & 4 & Roundabout with other cars \\ 
& & 5 & Lane merge with other cars \\ 
& & 6 & T-junction with other cars  \\ 
\bottomrule
\end{tabular}
\end{center}
\caption{Tasks in the Household and Driving Domains. Tasks with IDs 1 to 3 are generally perceived to be easier than tasks labelled with IDs 4 to 6.}
\label{fig:tasks}
\end{figure}

\subsection{Experimental Design}
\label{sec:method}
An overview of our experimental design is shown in Fig. \ref{fig:ExpDesign}. We explored three factors as independent variables: task category, task difficulty, and robot performance. Each independent variable consisted of two levels: two task categories, easy/difficult tasks, and robot success/failure. We used tasks in two domains, each with an appropriate robot (Fig. \ref{fig:tasks}): 
\begin{itemize}
	\item \textbf{Household}, which included two common categories of household tasks, i.e., picking and placing objects, and navigation in an indoor environment. The robot used was a real-world Fetch research robot with a 7-DOF arm, which performed \emph{live} demonstrations of the tasks in a lab environment that resembles a studio apartment. 
	\item \textbf{Driving}, where we used a Virtual Reality (VR) environment to simulate an autonomous vehicle (AV) performing tasks such as lane merging and parking, potentially with dynamic and static obstacles. Participants interacted with the simulation system via an Oculus Rift headset, which {provided a first-person viewpoint from the driver seat of the AV}. 
\end{itemize}
The robots were different in both settings and there were no cross-over tasks; in other words, the same experiment was conducted independently in each domain with the same protocol. Obtaining data from two separate experiments enabled us to discern if our hypotheses held in different contexts.

\begin{figure}
    \centering
    \begin{tabular}{c c}
        \centering
        \begin{subfigure}{.48\columnwidth}
            \centering
            \begin{tabular}{cc}
                \includegraphics[width=0.9\linewidth]{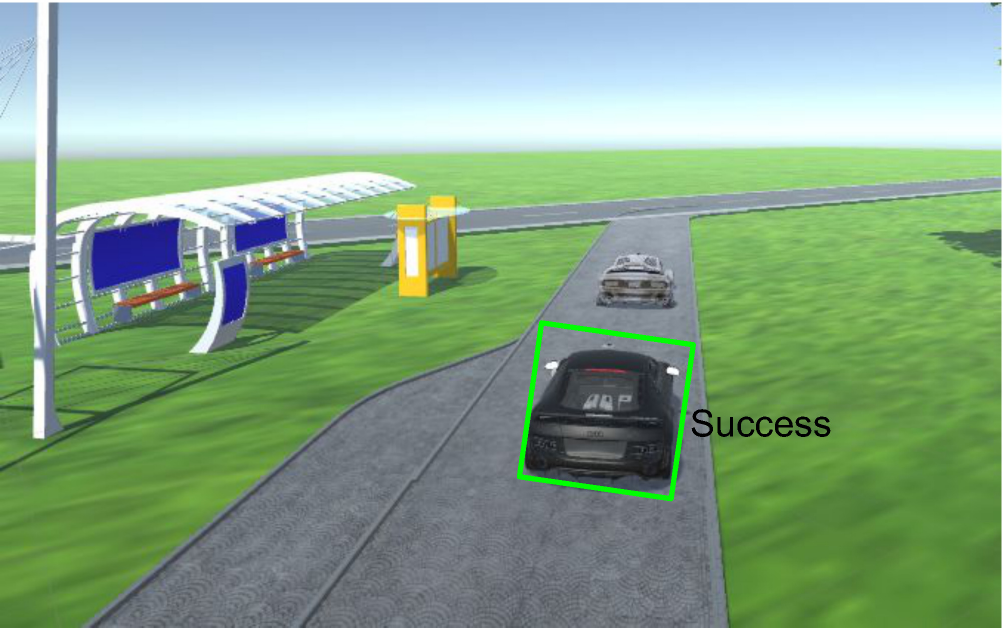}
                \\
                \\
                \includegraphics[width=0.9\linewidth]{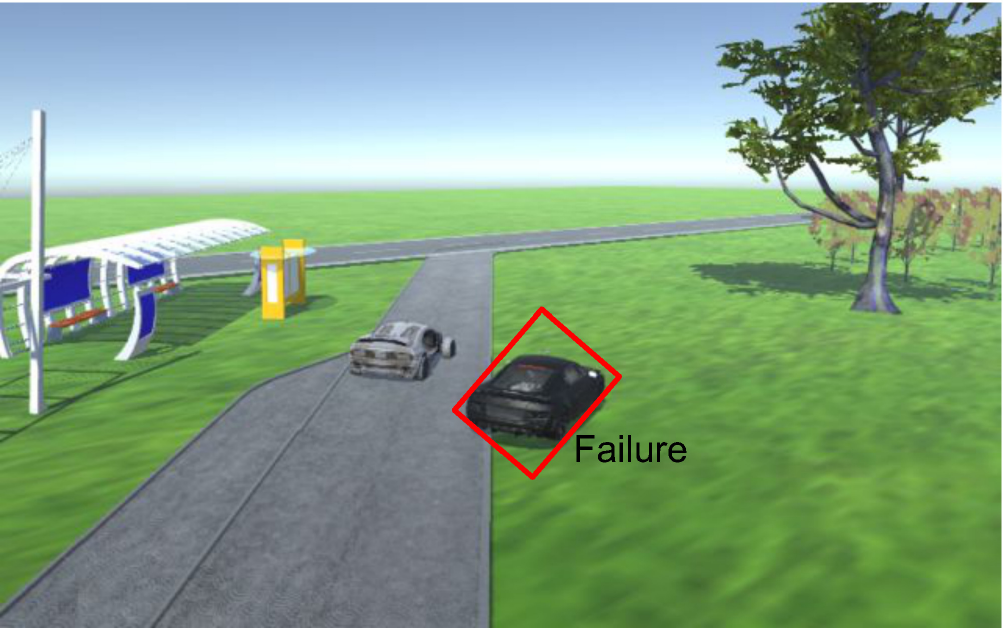}
            \end{tabular}
        \end{subfigure}

        \begin{subfigure}{.48\columnwidth}
            \centering
            \begin{tabular}{cc}
                \includegraphics[width=0.9\linewidth]{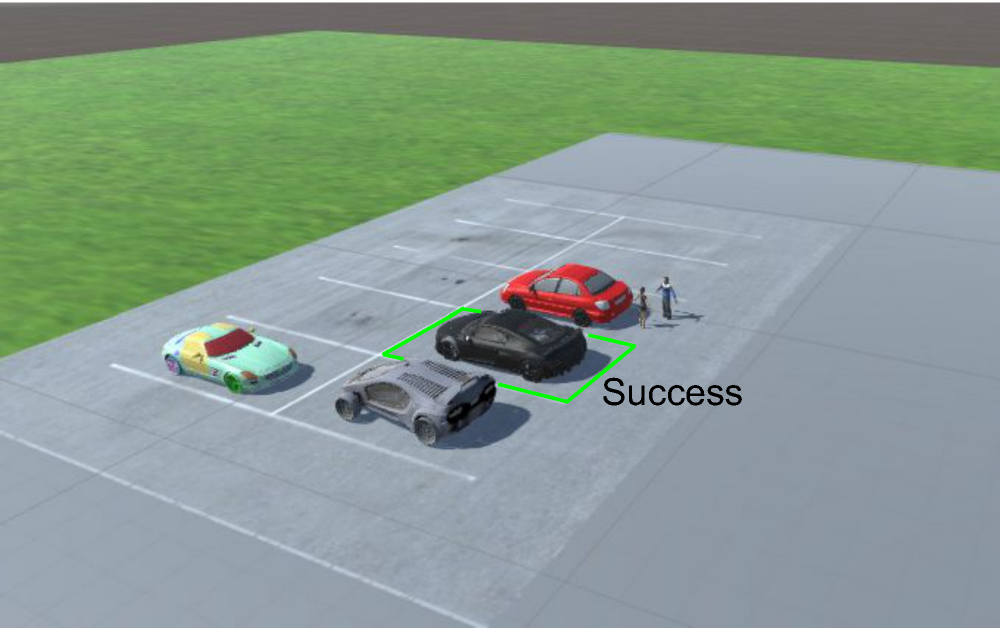}
                \\
                \\
                \includegraphics[width=0.9\linewidth]{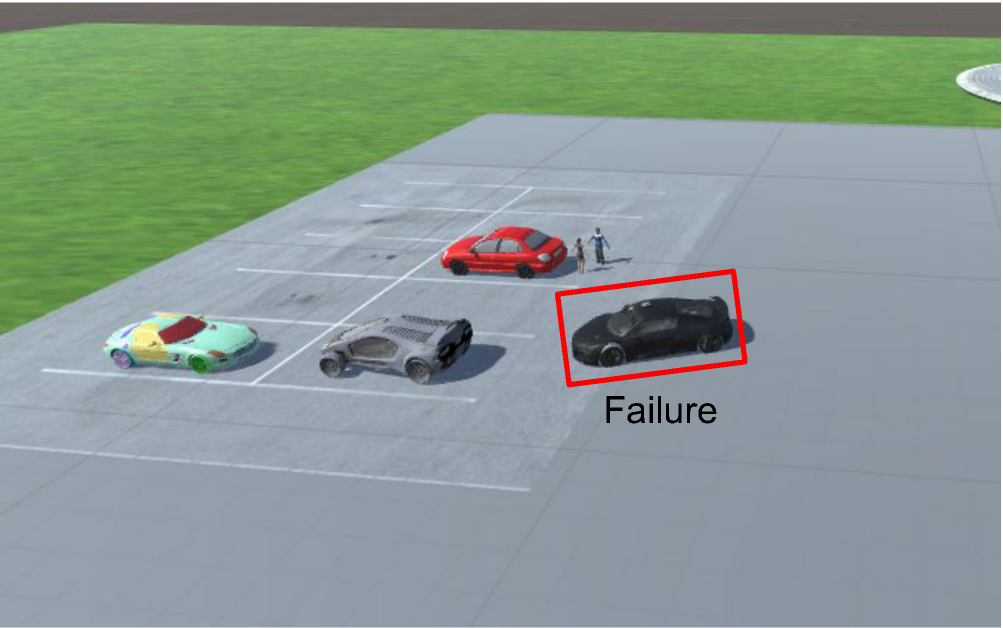}
            \end{tabular}
        \end{subfigure}
    \end{tabular}
    \caption{(Left Column) Autonomous car success (top) and failure (bottom) in the lane merge task. In the failure condition, the car drives off road and stops. (Right column) Autonomous car success (top) and failure (bottom) in the forwards parking task. In the failure condition, the car stops short of the parking spot.}
    \label{fig:variable-success-failure}
\end{figure}

In both domains, we developed pre-programmed success and failure demonstrations of robot performance {for all tasks}. ``Catastrophic'' failures were avoided to mitigate complete distrust of the robot; for the household navigation tasks, the robot was programmed to fail by moving to the wrong location. For picking and placing, the robot failed to grasp the target object. The autonomous car failed to park by stopping too far ahead of the lot, and failed to navigate (e.g., lane merge) by driving off the road and stopping (Fig. \ref{fig:variable-success-failure}). %

The primary dependent variables were the participants' subjective trust in the robot $a$'s capability to perform specific tasks. Participants indicated their degree of trust {given robot $a$} and task $x$ at time $t$, denoted as $\trust_{x,t}$, %
via a 7-point Likert scale in response to the agreement question: ``\emph{The robot is going to perform the task} [$x$]. \emph{I trust that the robot can perform the task successfully}''. In our form, the left-most point (1) indicated ``\emph{Strongly Disagree}'' and the right-most point (7) indicated ``\emph{Strongly Agree}''. 
From these task-dependent trust scores, we computed two derivative scores:
\begin{itemize}
    \item \textbf{Trust distance across tasks} $d_{\tau,t}(x,x') = |\trust_{x,t}-\trust_{x',t}|$, i.e., the 1-norm distance between scores for $x$ and $x'$ at time $t$.
    \item \textbf{Trust change over time} $\Delta \trust_x(t_1, t_2) = |\trust_{x,t_1}-\trust_{x,t_2}|$, i.e., the 1-norm distance between the scores for $x$ at $t_1$ and $t_2$.
\end{itemize}
As a general measure of trust, participants were also asked to complete Muir's questionnaire~\citep{muir1994trust, muir1996trust} pre-and-post exposure to the robot demonstrations. We also asked the participants to provide free-text justifications for their trust scores. 

\subsection{Robot Systems Setup}
For both the Fetch Robot and Autonomous Driving simulator, we developed our experimental platforms using the Robot Operating System (ROS). On the Fetch robot, we used the MoveIt motion planning framework and the Open Motion Planning Library~\citep{sucan2012} to pick and place objects, and the ROS Navigation stack for navigation in indoor environments. 

The VR simulation platform was developed using the Unity 3D engine. Control of the autonomous vehicle was achieved using the hybrid A* search algorithm~\citep{dolgov2010path} and a proportional-integral-derivative (PID) controller.

\subsection{Study Procedure}
We recruited 32 individuals (Mean age: 24.09 years, $SD = 2.37$, 46\% female) through an online advertisement and printed flyers on a university campus. {Experiments were conducted in our lab where participants were shown live demonstrations of the Fetch robot performing the tasks, or observed the AV's behavior using the driving simulator}. After signing a consent form and providing standard demographic data, participants were introduced to the robot. {Specifically, they were provided information about the robot's parts and basic functions, and then asked questions to ensure that they were paying attention and understood the information.} They then continued with the experiment's four stages:

\begin{enumerate}
	\item\textbf{Category and Difficulty Grouping:} To gain better control of the factors, participants were asked to group the 12 tasks evenly into the four cells shown in Fig. \ref{fig:ExpDesign}. As such, chosen observations matched a participant's own prior estimations. {We found that participant groupings were consistent --- the same grouping was observed across participants --- but there was individual differences within each difficulty group, e.g., some participants thought picking and placing a lemon was comparatively more difficult than a glass cup.}   
	\item\textbf{Pre-Observation Questionnaire:} Participants were asked to indicate their subjective trust on the three tested tasks using the measure instruments described above. 
	\item\textbf{Observation of Robot Performance:} Participants were randomly assigned to observe two tasks from a specific category and difficulty, and were asked to indicate their trust if the robot were to repeat the observed task. {The revised trust score is the baseline from which we evaluate the trust distance to tested tasks.}
	\item\textbf{Post-Observation Questionnaire and Debrief:} Finally, participants were asked to re-indicate their subjective trust on the three tested tasks, answered attention/consistency check questions\footnote{{Participants were asked several questions (e.g., what the last survey question was regarding) and to indicate their initial stated category/difficulty for a subset of the tasks.}}, and underwent a short de-briefing. 
\end{enumerate}

\subsection{Results}
\label{sec:results}

{In the following, we first report our primary findings using the task-dependent trust distance and change scores defined above. Then, we discuss the relationship between task-dependent trust and general trust (via comparison to the scores obtained from Muir's questionnaire).} For the driving domain, one participant's results were removed due to a failure to pass attention/consistency check questions. 

{For the Household domain, an ANOVA showed that the effect of the category group on the trust distance was significant, $F(1, 89) = 24.22, p < 10^{-5}$, as was the effect of difficulty, $F(1,89) = 14.05, p < 0.001$. The interaction between these two factors was moderately significant, $F(1,89) =  2.94, p = 0.089$. We also found a significant effect of category on trust change $F(1, 89) = 24.89, p < 10^{-6}$, and of success/failure outcomes on trust change $F(1, 89) = 10.04, p = 0.002$. Similar results were found for the driving domain.} 

\begin{figure}
\centering
\includegraphics[width=0.4\columnwidth]{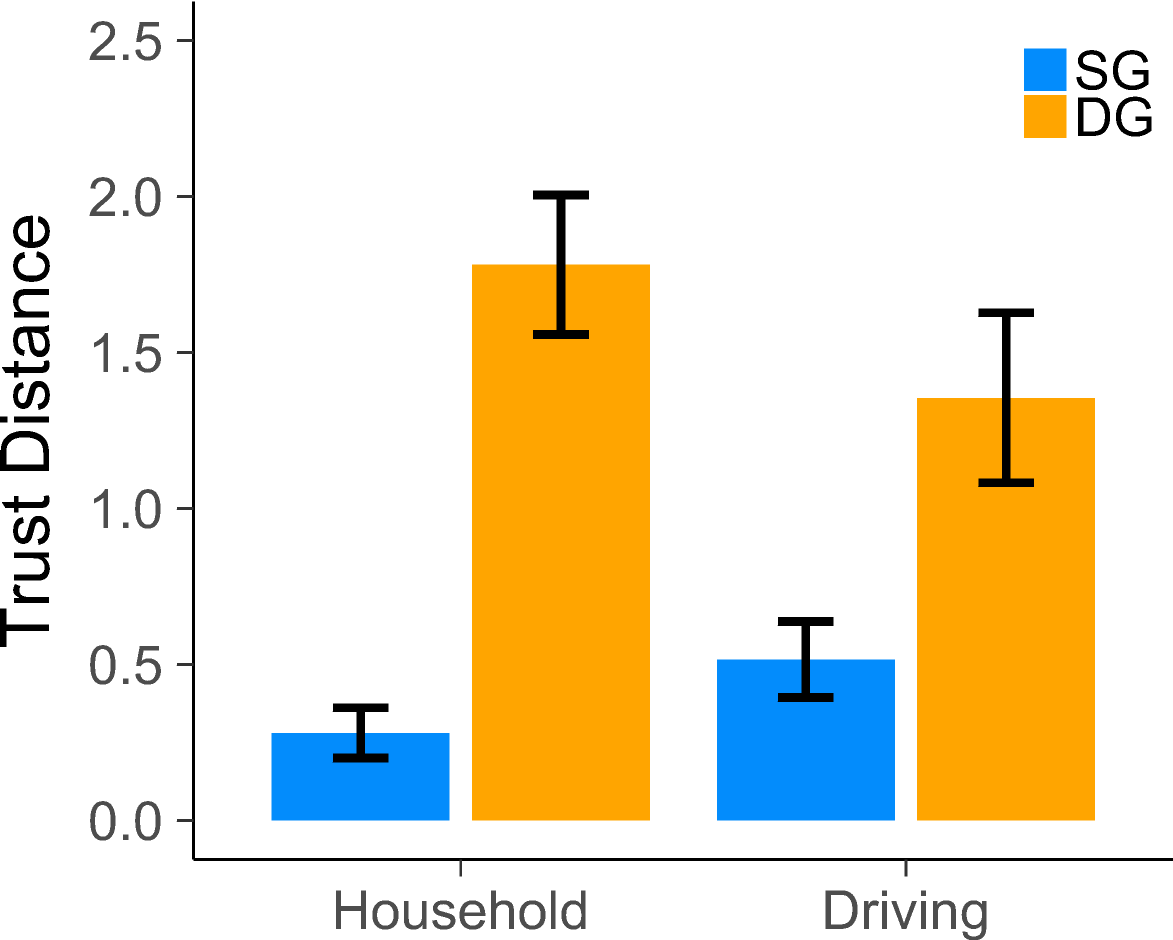}
	\caption{Trust distance between a given task and tasks in the same category group (SG) compared to tasks in a different category (DG). Trust in robot capabilities was significantly more similar for tasks in the same group.}
	\label{fig:H1}
\end{figure}

Fig. \ref{fig:H1} clearly shows that tasks in the same category (SG) shared similar scores (supporting \textbf{H1}); the {post-observation trust distances (from the observed task to the tested tasks)} are significantly lower $(M = 0.28, SE = 0.081)$ compared to tasks in other categories (DG) $(M = 1.78, SE = 0.22)$, $t(31) = -5.82$, $p < 10^{-5} $ for the household tasks. Similar statistically significant differences are observed for the driving domain, $t(30) = -2.755$, $p < 10^{-2}$.  {For both domains, we observe moderate effect sizes ($\approx 1$ on a Likert scale of 7), which suggests practical significance; the relative difference in trust may potentially affect subsequent decisions to delegate tasks to the robot.} 

\begin{figure}
\centering
\includegraphics[width=0.40\columnwidth]{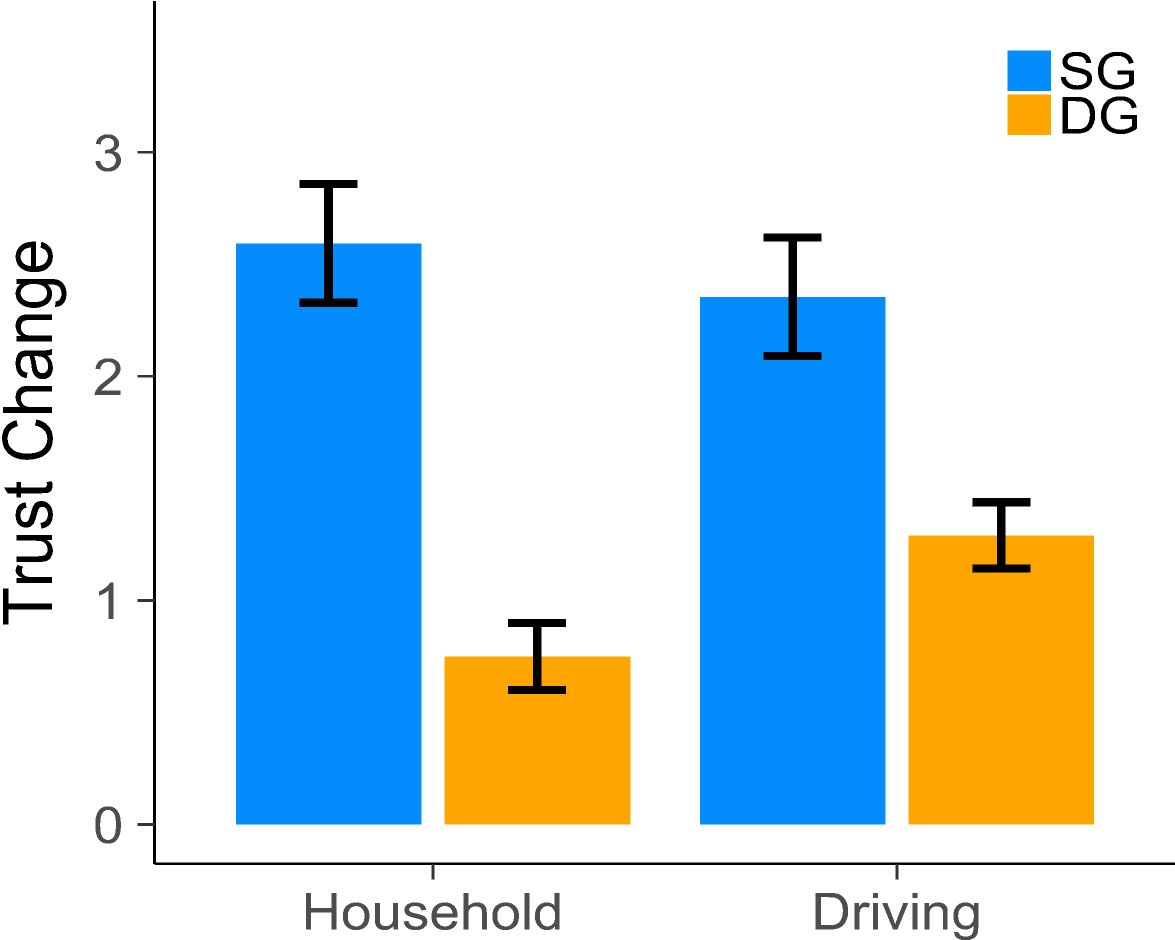}
	\caption{Trust change due to observations of robot performance. Trust increased (or decreased) significantly more for the {tested tasks in the same group (SG) as the observed task versus tasks in different groups (DG).}}
	\label{fig:H2}
\end{figure}

Fig. \ref{fig:H2} shows that the \emph{change} in human trust due to performance observations of a given task was moderated by the perceived similarity of the tasks (\textbf{H2}). The trust change {(between the pre-observation and post-observation trust scores for the three tested tasks)} is significantly greater for {tested tasks in the same group as the observed task (SG) than tasks in a different group (DG)}; $t(31) = 6.25$, $p < 10^{-6}$ for household and $t(30) = 3.46$, $p < 10^{-2}$ for driving. Note also that the trust change for DG was non-zero (one-sample $t$-test, $p < 10^{-2}$ across both domains for successes and failures), indicating that trust transfers even between task categories, albeit to a lesser extent. {Similar to the trust distances, the trust change effect sizes were moderately large indicating practical significance}.

\begin{figure*}
	\centering 
	\includegraphics[width=0.70\textwidth]{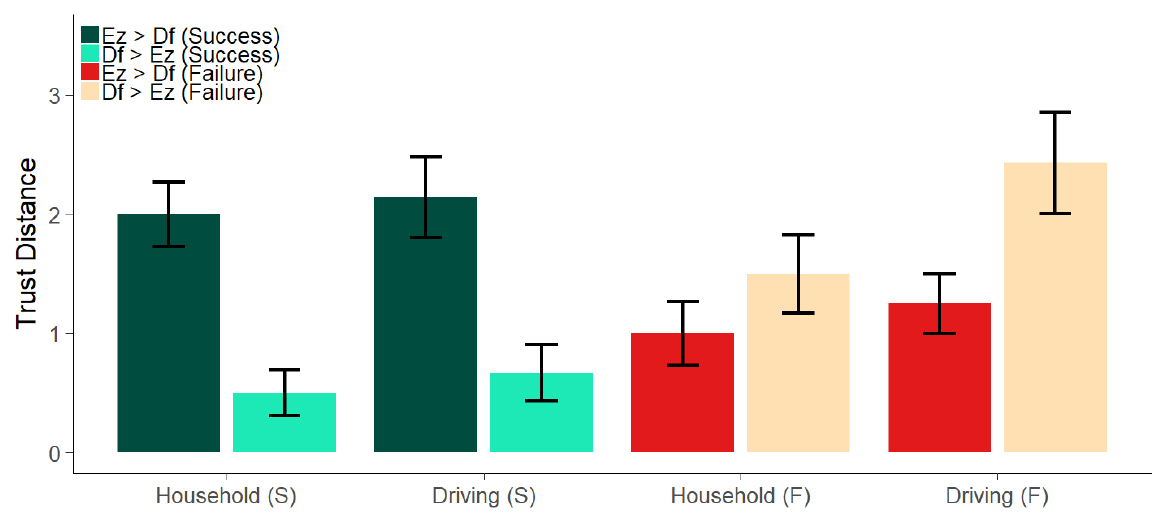}
    \caption{Trust distance between the observed task and a more difficult task (Ez $\rightarrow$ Df) against  when generalizing to a simpler task (Df $\rightarrow$ Ez). Participants who observed successful demonstrations of a difficult task trusted the robot to perform simpler tasks, but not vice-versa. 
    }
	\label{fig:H3}
\end{figure*}

We analyzed the relationship between perceived difficulty and trust transfer (\textbf{H3}) by first splitting the data into two conditions: participants who received successful demonstrations, and those that observed failures (Fig. \ref{fig:H3}). For the success condition, the trust distance among the household tasks was significantly less for tasks perceived to be easier than the observed task $(M = 2.0, SE = 0.27)$, compared to tasks that were perceived to be more difficult $(M = 0.5, SE = 0.27)$,  $t(14) = 4.58$, $p < 10^{-3}$. The hypothesis also holds in the driving domain, $M = 1.25$ $(SE = 0.25)$ v.s. $M = 2.43$ $(SE = 0.42)$, $t(14)=3.6827$, $p < 10^{-3}$.	For the failure condition (\textbf{H4}), the results were not statistically significant at the $\alpha=1\%$ level, but suggest that the effect was reversed; belief in robot \emph{inability} would transfer more to difficult tasks compared to simpler tasks.

\begin{figure}
	\centering
	\includegraphics[width=0.7\textwidth]{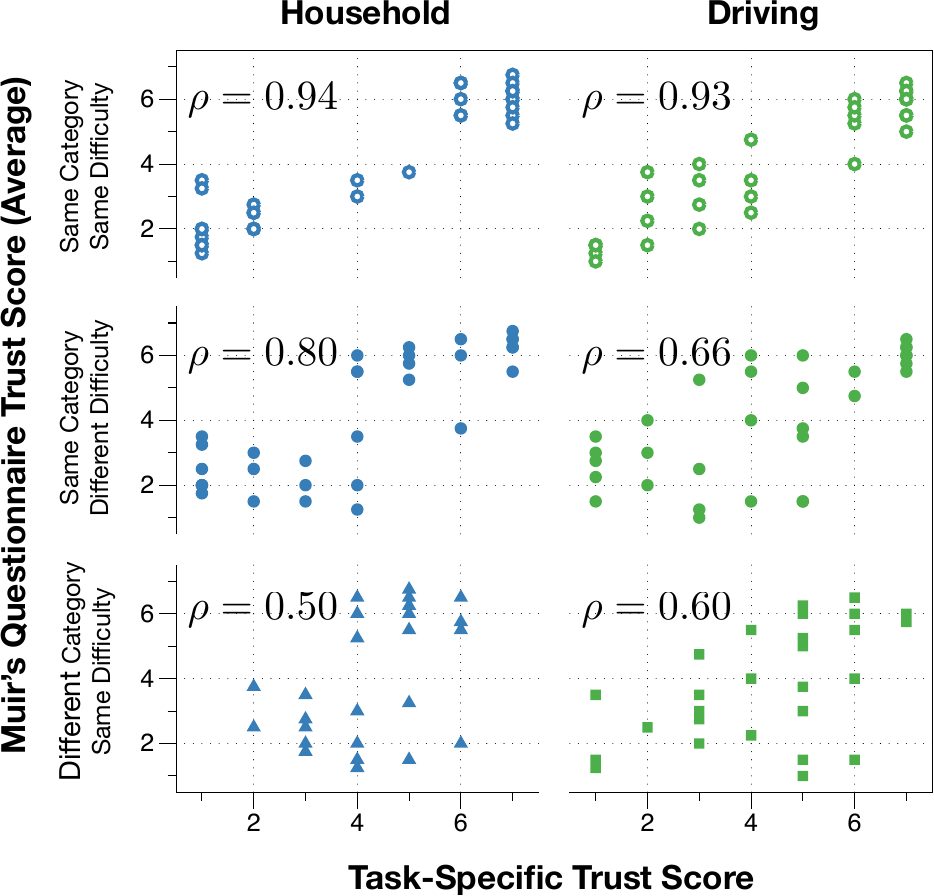}
	\caption{\cite{muir1994trust}'s Trust Score vs Task-specific trust scores (for tested tasks). The scores are positively correlated across the three task types, but with different strengths; the general measure is less predictive of task-specific trust for tasks in different categories (Pearson correlation, $\rho = 0.5-0.6$) compared to tasks with same category and difficulty ($\rho=0.93-0.94$).}
	\label{fig:Muir}
	\vspace{-0.5em}
\end{figure}

Thus far, we have focussed on task-specific trust; a key question is how this task-dependent trust differs from a ``general'' notion of trust in the robot as measured by Muir's questionnaire. Fig. \ref{fig:Muir} sheds light on this question; overall, task-specific and general trust are positively correlated but the degree of correlation depends greatly on the similarity of the task to previous observations. In other words, while general trust is predictive of task-specific trust, it does not capture the range or variability of human trust across multiple tasks.

\begin{table*}
\begin{center}
\centering \caption {Effects of Participant Characteristics on Initial Trust and Trust Dynamics. Un-corrected p-values are shown. {Bolded rows correspond to statistically significant coefficients at $\alpha = 0.1$ after Holm-correction.}}
\begin{adjustbox}{width=0.90\textwidth}
\begin{tabular}{l c c c c c c c c}
\toprule
\textbf{} & \multicolumn{4}{c}{\textbf{Initial Trust}} & \multicolumn{4}{c}{\textbf{Trust Change}}\\
\textbf{} & \textbf{coeff.} & \textbf{std err} &  \textbf{$t$-value} & \textbf{$\boldsymbol{\textrm{Pr}(>|t|)}$} & \textbf{coeff.} & \textbf{std err} &  \textbf{$t$-value} & \textbf{$\boldsymbol{\textrm{Pr}(>|t|)}$} \\
\midrule 
{Gender}& 0.43569 & 0.34425 & 1.266&0.211 & -0.17572 & 0.10067 & -1.745&0.086\\
{Computer Usage}\\
{$\qquad$Linear Term}&-0.57557& 0.31259 &-1.841& {0.071} &{0.20846}& {0.09142} &{2.280}& {0.026}\\
{$\qquad$Quadratic Term}&  0.46851 & 0.28246&1.659&0.103 &  0.06314 & 0.08261& 0.764&0.448\\
{$\qquad$Cubic Term}& 0.52029 &0.27504 &1.892 & {0.064} & -0.07554 &0.08043 &-0.939 &0.352 \\
{Experience with Robots} \\
{$\qquad$Linear Term}& \textbf{1.02462} & \textbf{0.35543}&\textbf{2.883}& \textbf{0.006} & -0.14597 & 0.10394&-1.404& 0.166\\
{$\qquad$Quadratic Term}& 0.06946 & 0.30585 & 0.227&0.821 & -0.05689 & 0.08944 & -0.636&0.527\\
{$\qquad$Cubic Term}& 0.27462 & 0.22962 &1.196&0.237 & 0.03136 & 0.06715 &0.467&0.642\\
{Experience with Video Games}\\
{$\qquad$Linear Term}& -0.54237& 0.44622&-1.215 & 0.223 & -0.23355& 0.13049&-1.790 & {0.079}\\
{$\qquad$Quadratic Term}& 0.18745& 0.32895&0.570 & 0.571 & 0.06359& 0.09620&0.661 & 0.511\\
{$\qquad$Cubic Term}& -0.07300& 0.26911&-0.271 & 0.787 & -0.03880& 0.07870&0.493 & 0.624\\
{$\qquad$Quartic Term}& -{0.61713}& {0.23753}&{-2.598} & {0.012} & \textbf{0.19424}& \textbf{0.06946}&\textbf{2.796} & \textbf{0.007}\\
\bottomrule
\end{tabular}
\end{adjustbox}
\label{tbl:regression}
\end{center}
\end{table*}

\subsection{Summary of Findings and Discussion}
Our main findings support the intuition that human trust transfers across tasks, but to different degrees. More specifically, similar tasks are more likely to share a similar level of trust (\textbf{H1}). Observations of robot performance changes trust both in the observed task, and also in similar yet unseen tasks (\textbf{H2}). Finally, trust transfer is asymmetric: positive trust transfers more easily to simpler tasks than to more difficult tasks (\textbf{H3}). These findings suggest that to infer human trust accurately in a collaboration across multiple tasks, robots should consider the similarity and difficulty of previous tasks.\par
  
\subsubsection{Qualitative Analyses.} Participant justifications for their trust scores were found to be consistent with the above findings. For example, a participant who was previously shown the robot successfully pick and place a plastic bottle and asked about her trust in the robot to pick and place a can, stated ``\emph{I trust this robot because the shape of the can of chips is similar to the bottle of soda}'', whilst another participant who observed failures stated he distrusted the robot because the task was ``\emph{highly similar to the last two failed tasks}''. 

The justifications also revealed that some participants were more willing to trust initially (higher dispositional trust), e.g., ``\emph{Yes, I first gave the robot the benefit of the doubt on a task I saw that similar robot can perform some of the time; then I revised my trust completely based on what it actually did on a similar task}''. Differences in perceived task difficulty also played a role in initial trust, ``\emph{I trust the robot because this seems like a simple enough task.}'' and in trust transfer, for example, ``\emph{Robot failed much easier task of navigating around a stationary item, so I don't think it can follow a moving object}''.

\subsubsection{Trust and Participant Characteristics.}

Finally, we examine how participant characteristics may affect dispositional and situational trust. Specifically, we analyzed the effect of four independent variables---gender, amount of computer usage, prior experience with video games, and prior experience  with robots---on the average initial trust and the average trust change. All four independent variables were self-reported; participants indicated their level of computer usage per week by selecting one of the following five choices: $<$10 hours, 10-20 hours, 20-30 hours, 30-40 hours, $>$40 hours. Prior experience with video games and robots was measured using the agreement questions, ``\emph{I'm experienced with video games}'' and ``\emph{I'm experienced with robots}'' using a 5-point Likert scale. We used the scores collected using Muir's questionnaire to compute the average initial trust and trust change. {A polynomial contrast model}~\citep{Saville1991} {was applied since the independent variables are ordinal and the true metric intervals between the levels are unknown\footnote{Polynomial contrast allows ordinal variables to enter not only linearly but also with higher order to better ascertain monotonic effects.}.} We also ran tests against the task-specific trust but the results were not significant; this was potentially due to participants being exposed to very different tasks.

Table \ref{tbl:regression} summarizes our results. {After correcting for family-wise error, we found a moderately significant association ($\alpha = 0.1$) between initial trust and prior experience with robots. Participants who had prior exposure to robots were more likely to trust the robots in our experiments. More experience with video games is significantly associated with  trust changes. Although not statistically significant, it may also be negatively associated with initial trust (Holm-corrected $p = 0.12$). These results suggest that participant characteristics, such as their prior experience with technology, do play a role in trust formation and dynamics. However, the relationships do not appear straightforward and we leave further examination of these factors to future work. }

\subsubsection{Study Limitations.} {In this work, each participant only observed the robot performing two tasks; we plan to investigate longer interactions involving multiple trust updates in future work. Furthermore, our reported results are based on subjective self-assessments in non-critical tasks. We believe our results to remain valid  when the robot's actions affect the human participant's goals. Our recent work}~\citep{Xie2019Hri,Chen2018Hri} {includes behavioral measures, such as operator take-overs and greater forms of risk (e.g., in the form of performance bonuses/penalties; these experiments also provide evidence for trust transfer across tasks.}.

\section{Computational Models for Trust Across Multiple Tasks}
\label{sec:ComputationalModels}

The results from our human subjects study indicate that trust is relatively rich mental construct. Perceived similarity between tasks plays a crucial role in determining trust transfer. Although we consider trust to be a useful information processing ``bottleneck'' in that it summarizes past experience with the robot, it does appear to be task-specific and hence, is more than a simple scalar quantity~\citep[as assumed in prior work][]{Chen2018Hri,Xu2016}. 

In this section, we present a richer model where trust is a \emph{task-dependent latent dynamic function} $\tau^\robot_t(\mathbf{x}): \mathbb{R}^d \rightarrow [0,1]$ that maps task features, $\mathbf{x}$, to {the continuous interval $[0,1]$} indicating trustworthiness of the robot to perform the task. {We assume that the task features are given and are sufficiently informative of the underlying tasks; for example, our experiments utilized word-vector features derived from English-language task descriptions, but visual features extracted from images or structured task descriptions may also be used.} 

This functional view of trust enables us to naturally capture trust differences across tasks, and can be extended to include other contexts; $\mathbf{x}$ can represent other factors, e.g., the current environment, robot characteristics, and observer properties. To model the dynamic nature of trust, we propose a Markovian function $g$ that updates trust, 
\begin{align}
	\tau^\robot_t = g(\tau^\robot_{t-1}, o^{\robot}_{t-1}) 
\end{align}
where $o^{\robot}_{t-1} = (\mathbf{x}_{t-1}, c^\robot_{t-1})$ is the observation of robot $\robot$ performing a task with features $\mathbf{x}_{t-1}$ at time $t-1$ with performance outcome $c^\robot_{t-1}$. The function $g$ serves to change trust given observations of robot performance, and as such, is a function over the space of trust functions. In this work, we consider binary outcomes $c^\robot_{t-1}\in \{+1, -1\}$ indicating success and failure respectively, but differences in performance can be directly accommodated via ``soft labels'' $c^\robot_{t-1}\in [-1, +1]$ without significant modifications to the presented methods.%

The principle challenge is then to determine appropriate forms for $\tau^\robot_t$ and $g$. In this work, we propose and evaluate three different approaches: (i) a Bayesian approach where we model a probability distribution over latent functions via a Gaussian process, (ii) a connectionist approach utilizing a recurrent neural network (RNN), and (iii) a hybrid approach that combines the aforementioned two methods. All three models leverage a learned \emph{psychological task space} in which similarities between tasks can  be efficiently computed. Furthermore, as differentiable models, they be trained using standard off-the-shelf gradient optimizers, and can be ``plugged'' into larger models or decision-making frameworks (e.g., deep networks trained via reinforcement learning).

\begin{figure*}
\centering
	\includegraphics[width=0.70\textwidth]{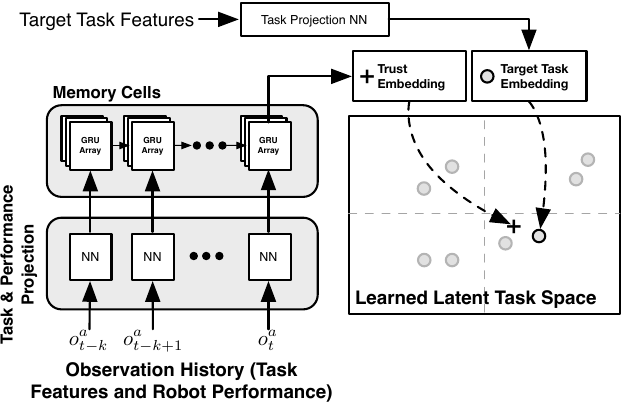}
	\caption{A High-level Schematic of the Neural Trust Model. The trust vector is updated using GRU cells as memory of previously observed robot performance. The model uses feed-forward neural networks to project tasks into a dot-product  space in which trust for a task can be efficiently computed.}
	\label{fig:NeuralTrustTransfer}
\end{figure*}

\subsection{Bayesian Gaussian Process Trust Model}
\label{sec:BayesianTrustModel}

In our first model, we view trust formation as a cognitive process, specifically, human function learning~\citep{Griffiths2009}. {We adopt a rational Bayesian framework, i.e., the human is learning about the robot capabilities by combining prior beliefs about the robot with evidence (observations of performance) via Bayes rule. More formally, let us denote the task at time $t$ as $\mathbf{x}_{t}$, and the robot $a$'s corresponding performance as $c^a_t$. Given binary success outcomes (where $c^\robot_t = 1$ indicates success), we introduce a latent function $f^a$ and model trust in the robot as,} 
\begin{align}
	\tau^a_t(\mat{x}_t) = & \int P(c^a_t = 1| f^a, \mat{x}_t)p_{t}(f^a)df^a 
\end{align}
{where $p_{t}(f^a)$ is the human's current belief over $f^a$, and $P(c^\robot_{t}|f^a, \mathbf{x}_{t})$ is the likelihood of observing the robot performance $c^\robot_{t}$ given the task $\mathbf{x}_{t}$. Intuitively, $f^a$ can be thought of as a latent ``unnormalized'' trust function that has range over the real number line. Given an observation of robot performance, the human's trust is updated via Bayes rule,}
\begin{align}
	p_{t}(f^a|\mathbf{x}_{t-1}, c^\robot_{t-1}) = \frac{P(c^\robot_{t-1}|f^a, \mathbf{x}_{t-1})p_{t-1}(f^a)}{\int P(c^\robot_{t-1}|f^a, \mathbf{x}_{t-1})p_{t-1}(f^a) df^a},
	\label{eq:bayesupdate}
\end{align}
{where $p_{t}$ is the posterior distribution over $f^a$.} %

{To use this model, we need to specify the prior $p_0(f^a)$  and likelihood $P(c^a_i| f^a, \mat{x})$ functions.} Similar to prior work in human function learning~\citep{Griffiths2009}, we place a Gaussian process (GP) prior over $f^a$,
\begin{align}
	p_{0}(f^a) = \mathcal{N}( m(\mat{x}), k(\mat{x}, \mat{x}')).
\end{align}
where $m(\cdot)$ is the prior mean function, and $k(\cdot, \cdot)$ is the kernel or covariance function. {The literature on GPs is large and we refer readers wanting more detail to} \cite{williams2006gaussian}. {In brief, a GP is a collection of random variables, of which any finite subset is jointly Gaussian}. {In this model, any given task feature $\mathbf{x}$ indexes a random variable representing the real function value $f^a(\mathbf{x})$ at specific location $\mathbf{x}$. The nice properties of Gaussians enable us to perform efficient marginalization, which makes the model especially attractive for predictive purposes. Note that the GP is completely parameterized by its mean and kernel functions, which we describe below.}
\paragraph{Covariance Function.} The kernel function is an essential ingredient for GPs and quantifies the similarities between inputs (tasks). {Popular kernel functions include the squared exponential and Mat\'{e}rn kernels} \citep{williams2006gaussian}. Although our task features are generally high dimensional (e.g., the word features used in our experiments), we consider tasks to live on a low-dimensional manifold, i.e., a psychological task space. With this in mind, we use a projection kernel:
\begin{align}
k(\mat{x}, \mat{x}') = \exp(-(\mat{x} - \mat{x}')^\top\mat{M}(\mat{x} - \mat{x}'))
\end{align}
with a low rank matrix $\mat{M} = \boldsymbol{\Lambda}\mat{L}\boldsymbol{\Lambda}^\top$ 
where $\boldsymbol{\Lambda} \in \mathbb{R}^{d \times k}$ and $\mat{L}$ is a diagonal matrix of length-scales capturing axis-aligned relevances in the projected task space. 

\paragraph{Capturing Prior Estimations of Task Difficulty and Initial Bias.} As our studies have shown, perceived difficulty results in an asymmetric transfer of trust (\textbf{H3}), which presents a difficulty for standard zero/constant-mean GPs given symmetric covariance functions. To address this issue, we explore two different approaches:
\begin{enumerate}
\item First, the mean function is a convenient way of incorporating a human's prior estimation of task difficulty; tasks which are presumed to be difficult (beyond the robot's capability) will have low values. Here, we have used a data-dependent linear function, $m(\mat{x}) = \boldsymbol{\beta}^\top\mat{x}$ where $\boldsymbol\beta$ is learned along with other GP parameters. 
\item A second approach is to use pseudo-observations $\mat{x}^+$ and $\mat{x}^-$ and associated $f^a$'s to bias the initial model. Intuitively, $\mat{x}^+$ ($\mat{x}^-$) summarizes the initial positive (negative) experiences that a person may have had. {The pseudo-observations are implemented simply as pre-observed data-points that the models are seeded with, prior to any trust updates.}  Similar to $\boldsymbol\beta$, these parameters are learned during training. {In our experiments, the pseudo-observations are trained using data from all the individuals in each training set, and thus, represent the ``average'' initial experience.}
\end{enumerate}
Both approaches allow the GP to accommodate the aforementioned asymmetry; the evidence has to counteract the prior mean function or pseudo-observations respectively. 

\paragraph{Observation Likelihood.} In standard regression tasks, the observed ``outputs'' are real-valued. However, participants in our experiments observed {binary  outcomes (the robot succeeded or failed)} and thus, we use the probit likelihood~\citep{neal1997monte},
\begin{align}
	P(c^\robot_{t}|f^a, \mathbf{x}_{t}) = \Phi \left( \frac{c^\robot_{t}(f^a(\mat{x}_{t}) - m(\mat{x}_t))}{\sigma_n^2} \right)
	\label{eq:likelihood}
\end{align}
where $\Phi(y) =  \frac{1}{\sqrt{2\pi}} \int_{-\infty}^y \exp\left(-\frac{t^2}{2}\right)dt$ is the CDF of the standard normal,
and $\sigma_n^2$ is the noise variance. {Here, $\Phi(y)$ is a response function that ``squashes'' the function value $y = f^a(\mat{x}) \in (-\infty, \infty)$ onto the range $[0,1]$. Alternative likelihoods can be used without changing the overall framework.}

\vspace{0.5em} %
\paragraph{Trust Updates via Approximate Bayesian Inference.} Unfortunately, the Bayesian update (\ref{eq:bayesupdate}) under the probit likelihood is intractable and yields a posterior process that is non-Gaussian. To address this problem and enable iterative trust updates, we employ approximate Bayesian inference: the posterior process is projected onto the closest GP as measured by the Kullback-Leibler divergence, $\mathrm{KL}(p_t||q)$, and $q$ is our GP approximation~\citep{csato2002a}. Minimizing the KL divergence is equivalent to matching the first two moments of $p_t$ and $q$, which can be performed analytically. The update equations in their natural parameterization forms are given by:
\begin{align}
	\mu_t(\mat{x}) & =  \boldsymbol{\alpha}_t^\top\mathbf{k}(\mat{x}) \\
	k_t(\mat{x},\mat{x}') & =  k(\mat{x},\mat{x}') + \mathbf{k}(\mat{x})^\top\mathbf{C}_t\mathbf{k}(\mat{x}')
	\label{eq:updatek}
\end{align}
where $\boldsymbol{\alpha}$ vector and $\mathbf{C}$ are updated using:
\begin{align}
	\boldsymbol{\alpha}_{t} & = \boldsymbol{\alpha}_{t-1} + \gamma_1(\mathbf{C}_{t-1} \mathbf{k}_{t} + \mathbf{e}_{t}) \label{eq:updatealpha}\\
	\mathbf{C}_{t} & = \mathbf{C}_{t-1} + \gamma_2(\mathbf{C}_{t-1} \mathbf{k}_{t} + \mathbf{e}_{t})(\mathbf{C}_{t-1} \mathbf{k}_{t} + \mathbf{e}_{t})^{\top}  \label{eq:updateC}
\end{align}
where $\mathbf{k}_{t} = [k(\mat{x}_{1}, \mat{x}_{t}), ..., k(\mat{x}_{t-1}, \mat{x}_{t})]$, $\mathbf{e}_{t}$ is the $t^{\mathrm{th}}$ unit vector and the scalar coefficients $b_1$ and $b_2$ are given by:
\begin{align}
	\gamma_1 & = \partial_{f^a} \log \int P(c^\robot_{t}|f^a, \mathbf{x}_{t}) df^a = \frac{c^a_i \partial\Phi}{\sigma_x \Phi}\\
	\gamma_2 & = \partial^2_{f^a}  \log \int P(c^\robot_{t}|f^a, \mathbf{x}_{t}) df^a = \frac{1}{\sigma_x^2}\left[ \frac{\partial^2\Phi}{\Phi} - \left( \frac{\partial\Phi}{\Phi} \right) \right] 
\end{align}
where $\partial\Phi$ and $\partial^2\Phi$ are the first and second derivatives of $\Phi$ evaluated at $\frac{c^a_i (\mu_t(\mat{x}) - m(\mat{x}))}{\sigma_\mat{x}}$. 

\paragraph{Trust Predictions.} Given (\ref{eq:updatealpha}) and (\ref{eq:updateC}), predictions can be made with the probit likelihood (\ref{eq:likelihood}) in closed-form: 
\begin{align}
	\tau^a_t(\mat{x}) = & \int P(c^a = 1| f^a, \mat{x})p_{t}(f^a)df^a \nonumber \\
 = & \Phi\left(\frac{\mu_t(\mat{x}) - m(\mat{x})}{\sigma_\mat{x}}\right)
\end{align}
where $\sigma_\mat{x} = \sqrt{ \sigma_n^2 + k_t(\mat{x}_i,  \mat{x}_i)}$.

\subsection{Neural Trust Model} 
\label{sec:NeuralTrustModel}

The Gaussian process trust model is based on the assumption that human trust is essentially Bayesian in nature. However, this assumption may be too restrictive since humans are not thought to be fully rational or Bayesian\footnote{Moreover, whether brains are truly Bayesian remains a matter of debate within the cognitive sciences~\citep{bowers2012bayesian}.}. Here we consider an alternative ``data-driven'' approach based on recent advances in deep neural models. 

 The architecture of our neural trust model is illustrated in Fig. \ref{fig:NeuralTrustTransfer}. We leverage a learned task representation or ``embedding'' space $Z \subseteq \mathbb{R}^k$ and model trust as a parameterized function over this space. {The key idea is that (un-normalized) trust for a task is obtained via an inner-product between two components: a trust-vector $\btheta_t$ and a task representation $\mathbf{z}$. The trust vector $\btheta_t$ is a compressed representation of the human's prior interaction history (observations of robot performance $o^a_t$) and is derived using a recurrent neural network. The task representations $\mathbf{z}$ are derived using a learned function $f_z(\mathbf{x})$ over task features $\mathbf{x}$. To obtain a normalized trust score between $[0,1]$, we use the sigmoid function,} 
\begin{align}
	\tau^a_t(\mathbf{x}; \btheta_t) = \mathrm{sigm}(\btheta_t^\top f_z(\mathbf{x})) = \mathrm{sigm}(\btheta_t^\top\mathbf{z}).
\end{align}
The trust function $\tau^a_t$ is fully parameterized by $\btheta_t$ and its linear form has benefits: it is efficient to compute given a task representation $\taskz$ and is interpretable in that the latent task space $Z$ can be examined, similar to other dot-product spaces, e.g., word embeddings~\citep{mikolov2013}. Similar to the GP, $Z$ can be seen as a psychological task space in which the similarities between tasks can be easily ascertained.

\paragraph{Task Projection.} Whilst it is possible to train the model to learn separate task representations $\mathbf{z}$ for each task in the training set, this approach limits the model to only seen tasks. Our aim was to create a general model that potentially generalizes to new tasks. One could use the task features $\mathbf{x}$ directly in the trust function, but there is no guarantee that the task features would form a space in which dot products would give rise to meaningful trust scores. As such, we project observed task features $\taskfeat$ into $Z$ via a nonlinear function, specifically, a fully-connected neural network,  
\begin{align}
	\taskz &= f_z(\mathbf{x}) = \mathrm{NN}(\mat{x}; \theta_z)
	\label{eq:NNproj}
\end{align}
where $\theta_z$ is the set of network parameters. Similarly, the robot's performance $\performance $ is projected via a neural network, $\perfz = \mathrm{NN}( \performance ; \theta_\performance )$. During trust updates, both the task and performance representations are concatenated, $\hat{\taskz}_i = [ \taskz ; \perfz]$, as input to the RNN's memory cells.

\vspace{0.5em} %
\paragraph{Trust Updating via Memory Cells.} We model the trust update function $g$ using  a RNN with parameters $\theta_g$,
\begin{align}
	\btheta_t &= \mathrm{RNN}(\btheta_{t-1}, \hat{\mat{z}}_{t-1}; \theta_g).
	\label{eq:RNNupdate}
\end{align}
In this work, we leverage on the Gated Recurrent Unit (GRU)~\citep{cho2014}, which is a  variant of long short-term memory~\citep{Hochreiter1997} with strong empirical performance~\citep{jozefowicz2015empirical}. In brief, the GRU learns to control two internal ``gates''---the update and reset gates---that affect what it remembers and forgets. Intuitively, the previous hidden state is forgotten when the reset gate's value nears zero. As such, cells that have active reset gates have learnt to model short-term dependencies. In contrast, cells that have active update gates model long-term dependencies~\citep{cho2014}. Our model uses an array of GRU cells that embed the interaction history up to time $t$ as a ``memory state'' $\mat{h}_t$, which serves as our trust parameters $\btheta_t$.

More formally, a GRU cell $k$ that has state $h_{t-1}^{(k)}$ and receives a new input $\hat{\mat{z}}_t$, is updated via 
\begin{align}
	h_t^{(k)} = (1-v_t^{(k)})h^{(k)}_{t-1} + v_t^{(k)}\tilde{h}_{t}^{(k)},
\end{align}
i.e., an interpolation of its previous state and a candidate activation $\tilde{h}_{t}^{(k)}$. This interpolation is affected by the update gate $v_t^{(k)}$,  which is parameterized by matrices $\mathbf{W}_v$ and $\mathbf{U}_v$,
\begin{align}
	v_t^{(k)} = \textrm{sigm}\big([\mat{W}_v\hat{\mat{z}}_{t} + \mat{U}_v\mat{h}_{t-1})]_k\big).
\end{align}
The candidate activation $\tilde{h}_{t}^{(k)}$ is given by
\begin{align}
	\tilde{h}_{t}^{(k)} = \mathrm{tanh}([\mat{W}\hat{\mat{z}}_t + \mat{U}(\mat{r}_t \odot \mat{h}_{t-1})]_k )
\end{align}
where $\odot$ denotes element-wise multiplication. The reset gate $r^{(j)}_t = [\mat{r}_t]_k$ is parameterized by two matrices $\mat{W}_r$ and $\mat{U}_r$,
\begin{align}
	r^{(k)}_t = \textrm{sigm}\big([\mat{W}_r\hat{\mat{z}}_{t} + \mat{U}_r\mat{h}_{t-1})]_k\big)
\end{align}

\subsection{A GP-Neural Trust Model}
Both the neural and Bayesian models assume Markovian trust updates and that trust summarizes past experience with the robot. They differ principally in terms of the inherent flexibility of the trust updates. In the RNN model, the update parameters, i.e., the gate matrices, are learnt. As such, it is able to adapt trust updates to best fit the observed data. However, the resulting update equations do not lend themselves easily to interpretation. On the other hand, the GP employs fixed-form updates that are constrained by Bayes rule. While this can hamper predictive performance (humans are not thought to be fully Bayesian), the update is interpretable and may be more robust with limited data.

A natural question is whether we can formulate a ``structured'' trust update that combines the simplicity of the Bayes update, while allowing for some additional flexibility. Here, we examine a variant of the GP model that incorporates a neural component in the mean function update. In particular, we modify Eq. (\ref{eq:updatealpha}) with an additional term:
\begin{align}
\boldsymbol{\alpha}_{t} = & \boldsymbol{\alpha}_{t-1} + \gamma_1(\mathbf{C}_{t-1} \mathbf{k}_{t} + \mathbf{e}_{t}) +  u(\boldsymbol{\alpha}_{t-1}, \mathbf{C}_{t-1}\mathbf{k}_{t}, \boldsymbol{\Lambda}\mat{x}_{t-1}, c^\robot_{t-1})
\label{eq:bayesnnupdate}
\end{align}
where $u(\cdot)$ models any residual not fully captured by the Bayes update. The function $u$ takes as input the previous mean parameters $\boldsymbol{\alpha}_{t-1}$, $\mathbf{C}_{t-1}\mathbf{k}_{t}$, the latent task vector $\mathbf{z}_{t-1} = \boldsymbol{\Lambda}\mat{x}_{t-1}$, and the robot performance $c^\robot_{t-1}$. The key idea here is that a neural component may modify the posterior distribution in a data-driven (non-Bayesian) manner to better capture the intricacies of human trust updates. In our experiments, $u(\cdot)$ is a simple feed-forward neural network, but alternative models can be used without changing the overall framework. For example, using a GRU-based network would enable non-Markovian updates and may further improve performance.

\section{Experiments}
\label{sec:experiments}

Our experiments were designed to establish if the proposed trust models that incorporate inter-task structure outperform existing baseline methods. In particular, we sought to answer three questions:
\begin{itemize}
	\item [\textbf{Q1}] Is it necessary to model trust transfer, i.e., do the proposed function-based models perform better than existing approaches when tested on unseen participants? 
	\item [\textbf{Q2}] Do the models generalize to unseen tasks? 
	\item [\textbf{Q3}] Is it necessary to model differences in initial bias, specifically perceptions of task difficulty?  
	\item [\textbf{Q4}] Does incorporating additional flexibility into the GP trust updates improve performance?
\end{itemize}

\subsection{Experimental Setup}
To answer these questions, we conducted two separate experiments. \textbf{Experiment E1} was a variant of the standard 10-fold cross-validation where we held-out data from 10\% of the participants (3 people) as a test set. This allowed us to test each model's ability to generalize to unseen participants on the same tasks. To answer question \textbf{Q2}, we performed a leave-one-out test on the tasks (\textbf{Experiment E2}); we held-out all trust data associated with one task and trained on the remaining data. This process yielded 12 folds, one per task.

\vspace{0.5em} %
\paragraph{Trust Models.} We evaluated six models in our experiments:
\begin{itemize}
	\item \textbf{GP}: A constant-mean Gaussian process trust model;
	\item \textbf{PMGP}: The GP trust model with prior mean function;
    \item \textbf{POGP}: The GP trust model with prior pseudo-observations;
    \item \textbf{RNN}: The neural RNN trust model;
	\item \textbf{GPNN}: The Bayesian GP-neural trust model with prior pseudo-observations;
	\item \textbf{LG}: A linear Gaussian trust model similar to the updates used in OPTIMo~\citep{xu2015optimo};
	\item \textbf{CT}: A baseline model with constant trust.
\end{itemize} 
{The baseline CT and LG models did not utilize task features as they do not explicitly consider trust variation across tasks. The general LG model applies linear Gaussian updates}: 
\begin{align}
p(\trust_t | \trust_{t-1}, c^a_{t-1}, c^a_{t-2}) = \mathcal{N}\left(\mathbf{w}_{\textsc{LG}}^\top 
\begin{bmatrix}
 	\trust_{t-1} \\
 	 c^a_{t-1} \\
 	c^a_{t-1} - c^a_{t-2}
 \end{bmatrix}, \sigma_{\textsc{LG}}^2 \right)
 \end{align}
{where $\mathbf{w}_{\textsc{LG}}$ and $\sigma_{\textsc{LG}}^2$ are learned parameters. In our dataset, the robot $a$'s performance in the two time steps was the same (both success/failure). Hence, the updated trust only depends on the previous trust $\trust_{t-1}$ and robot performance $c^a_{t-1}$.}

We implemented all the models using the PyTorch framework~\citep{paszke2017automatic}. %
Preliminary cross-validation runs were conducted to find good parameters for the models. The RNN used a two layer fully-connected neural network with 15 hidden neurons and Tanh activation units to project tasks to a 30-dimensional latent task space (Eqn. (\ref{eq:NNproj})). The trust updates, Eqn. (\ref{eq:RNNupdate}), were performed using two layers of GRU cells. A smaller 3-dimensional latent task space was used for the GP models. GP parameters were optimized during learning, except the length-scales matrix, which was set to the identity matrix $\mat{L} = \mat{I}$; fixing $\mat{L}$ resulted in a smoother optimization process. For the GPNN, $u(\cdot)$ was set as a simple two-layer feed-forward neural network with 20 neurons-per-layer and Tanh activation units.

\paragraph{Datasets.} The models were trained using the data collected in our human subjects study. The RNN and GP-based models were \emph{not} given direct information about the difficulty and group of the tasks since this information is typically not known at test time. Instead, each task was associated with a 50-dimensional GloVe word vector~\citep{pennington2014glove} computed from the task descriptions in Fig. \ref{fig:tasks} (the average of all the word vectors in each description). {Complete task descriptions and code to derive the features are available in the online supplementary material}~\citep{SuppMat}.  

\paragraph{Training.} In these experiments, we predict how each individual's trust is dynamically updated. The tests are not conducted with a single ``monolithic'' trust model across all participants. Rather, training entails learning the latent task space and model parameters, which are shared among participants, e.g., $\boldsymbol\beta$ and $\boldsymbol{\Lambda}$ for the PMGP and the gate matrices for the GRU. However, each participant's model is updated \emph{only} with the tasks and outcomes that the participant observes. 

To learn these parameters, all models are trained ``end-to-end''. We applied maximum likelihood estimation (MLE) and optimized model parameters $\boldsymbol\theta$ using the Bernoulli likelihood of observing the normalized trust scores (as soft labels): 
\begin{align}
	\mathcal{L}(\theta) &= -\sum \hat{\tau}^a\log(1-\tau^a(\mathbf{x})) + (1-\hat{\tau}^a)\log(1-\tau^a(\mathbf{x}))
\end{align}
where $\hat{\tau}^a$ is the observed normalized trust score. In more general settings where trust is \emph{not} observed, the models can be trained using observed human actions, e.g., in~\citep{ChenMin2017}. We employed the \textsc{Adam} algorithm~\citep{kingma2014adam} for a maximum of 500 epochs, with early stopping using a validation set comprising 15\% of the training data. 

\paragraph{Evaluation} Evaluation is carried out on both pre-and-post-update trust. For both experiments, we computed two measures: the average Negative Log-likelihood (NLL) and Mean Absolute Error (MAE). However, we observed that these scores varied significantly across the folds (each participant/task split). To mitigate this effect, we also computed relative Difference from Best (DfB) scores: $
	\mathrm{NLL}_\mathrm{DfB}(i,k) = \mathrm{NLL}(i,k) - \mathrm{NLL}^*(i)$, 
where $\mathrm{NLL}(i,k)$ is the NLL achieved by model $k$ on fold $i$ and $\mathrm{NLL}^*(i)$ is the best score among the tested models on fold $i$. $\mathrm{MAE}_\mathrm{DfB}$ is similarly defined.

\subsection{Results}
\begin{figure}
	\centering 
	\includegraphics[width=0.50\columnwidth]{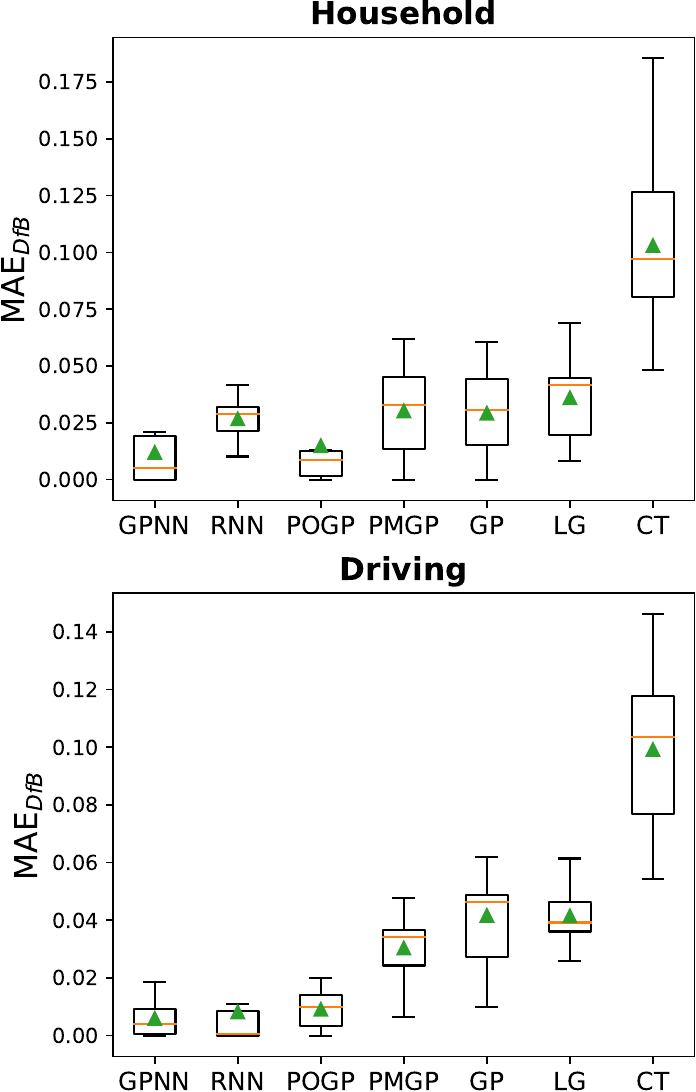}
    \caption{$\mathrm{MAE}_\mathrm{DfB}$ scores for experiment \textbf{E1} with medians (lines) and means (triangles) shown. The proposed task-dependent  trust models (GPNN, RNN, and POGP) models are superior at predicting trust  scores on unseen test participants. The GPNN achieves the lowest average $\mathrm{MAE}_\mathrm{DfB}$ scores across the two domains.\label{fig:partsplits}}
    \end{figure}

Results for \textbf{E1} are summarized in Tbl. \ref{tbl:E1} with boxplots of $\mathrm{MAE}_\mathrm{DfB}$ shown in Fig. \ref{fig:partsplits}. In brief, the GPNN, RNN,  and POGP outperform the other models on both datasets across the performance measures. The POGP makes better predictions on the Household dataset, whilst the RNN performed better on the Driving dataset. The GPNN, however, obtains good performance across the datasets. In addition, the GP achieves better or comparable scores on average relative to LG and CT. Taken together, these results indicate that the answer to \textbf{Q1} is in the affirmative: accounting for trust transfer between tasks leads to better trust predictions. 
    \begin{figure}
	\centering 
	\includegraphics[width=0.50\columnwidth]{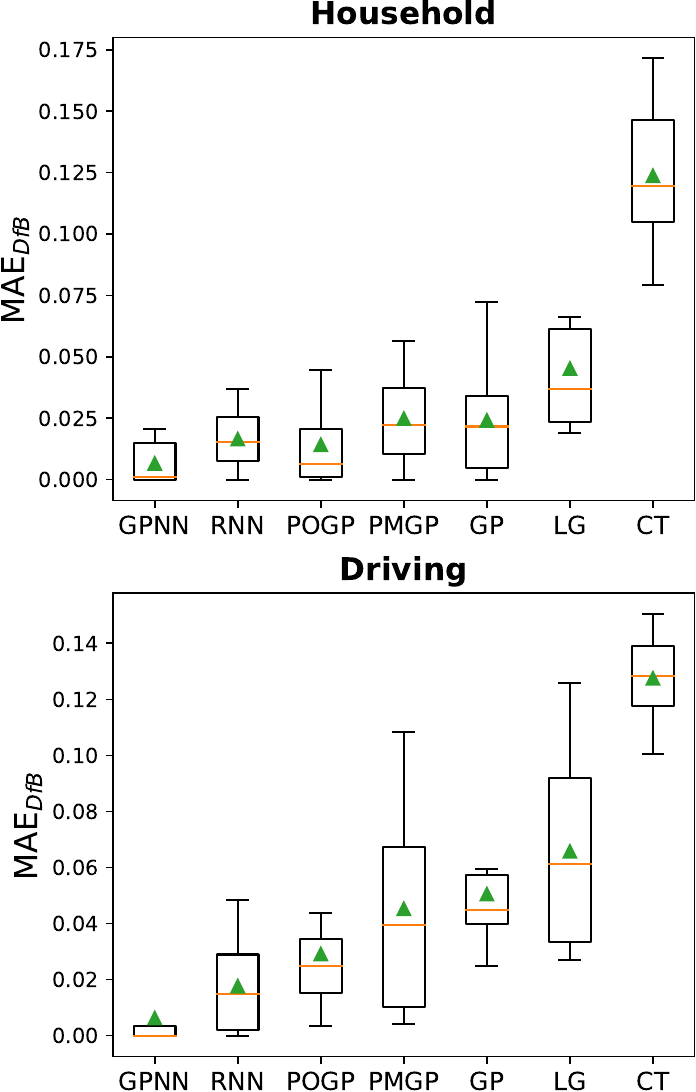}
    \caption{$\mathrm{MAE}_\mathrm{DfB}$ scores for experiment \textbf{E2} with medians (lines) and means (triangles) shown. The proposed task-dependent  trust models (GPNN, RNN, and POGP) models are superior at predicting trust  scores on unseen test tasks. Similar to Fig. \ref{fig:partsplits}, the GPNN achieves the lowest average $\mathrm{MAE}_\mathrm{DfB}$ scores across the two domains.   \label{fig:tasksplits}}
    \end{figure}

Next, we turn our attention to \textbf{E2}, which is potentially the more challenging experiment. The GPNN, RNN, and POGP again outperform the other models (see Tbl. \ref{tbl:E2} and Fig. \ref{fig:tasksplits}). Both models are able to make accurate trust predictions on \emph{unseen} tasks (\textbf{Q2}), suggesting that (i) the word vectors are informative of the task, and (ii) the models learnt reasonable projections into the task embedding space.

To answer \textbf{Q3} (whether modeling initial human bias is required), we examined the differences between the POGP, PMGP, and GP. The PO/PMGP achieved lower or similar scores to the GP model across the experiments and domains, indicating that difficulty modeling enabled better performance. The pseudo-observation technique POGP always outperformed the linear mean function approach PMGP, suggesting initial bias is nonlinearly related to the task features. Potentially, using a non-linear mean function may allow PMGP to achieve similar performance to POGP. 

Finally, we observed that the including additional flexibility in the GP mean update improved model performance (\textbf{Q4}). As stated above, the GPNN achieves similar or better performance on both datasets compared to  the other GP variants.

\begin{table}
\begin{center}
\caption{Model Performance on Held-out Participants (Experiment \textbf{E1}). Average Negative log-likelihood (NLL) and Mean Absolute Error (MAE) scores shown with standard errors in brackets. Best scores in \textbf{bold}.\label{tbl:E1}}
\begin{tabular}{ c c c c c  } 
\toprule
\multirow{2}{*} {Models} & \multicolumn{2}{c} {Household} & \multicolumn{2}{c} {Driving}\\
& NLL & MAE & NLL & MAE \\
\midrule
\multirow{2}{*} {GPNN} & \textbf{0.558}& \textbf{0.158}& 0.555 & \textbf{0.172}\\
& (\textbf{0.028})& (\textbf{0.011})& (0.026) & (\textbf{0.010})\\
\multirow{2}{*} {RNN} & 0.571 & 0.173 & \textbf{0.549}& 0.175 \\
& (0.023) & (0.010) & (\textbf{0.024})& (0.011) \\
\multirow{2}{*} {POGP} & 0.558 & 0.161 & 0.553 & 0.176 \\
& (0.027) & (0.013) & (0.025) & (0.012) \\
\multirow{2}{*} {PMGP} & 0.577 & 0.176 & 0.567 & 0.197 \\
& (0.019) & (0.010) & (0.018) & (0.011) \\
\multirow{2}{*} {GP} & 0.575 & 0.175 & 0.588 & 0.208 \\
& (0.023) & (0.013) & (0.022) & (0.012) \\
\multirow{2}{*} {LG} & 0.578 & 0.182 & 0.584 & 0.208 \\
& (0.023) & (0.011) & (0.022) & (0.011) \\
\multirow{2}{*} {CT} & 0.662 & 0.249 & 0.649 & 0.266 \\
& (0.029) & (0.016) & (0.017) & (0.010) \\
\bottomrule
\end{tabular}
\end{center}
\end{table}
\begin{table}
\begin{center}
\caption{Model Performance on Held-out Tasks (Experiment \textbf{E2}). Average Negative log-likelihood (NLL) and Mean Absolute Error (MAE) scores shown with standard errors in brackets. Best scores in \textbf{bold}. \label{tbl:E2}}
\begin{tabular}{ c c c c c  } 
\toprule
\multirow{2}{*} {Models} & \multicolumn{2}{c} {Household} & \multicolumn{2}{c} {Driving}\\
& NLL & MAE & NLL & MAE \\
\midrule
\multirow{2}{*} {GPNN} & \textbf{0.533}& \textbf{0.156}& 0.542 & \textbf{0.163}\\
& (\textbf{0.014})& (\textbf{0.007})& (0.019) & (\textbf{0.009})\\
\multirow{2}{*} {RNN} & 0.542 & 0.166 & \textbf{0.531}& 0.174 \\
& (0.012) & (0.006) & (\textbf{0.016})& (0.014) \\
\multirow{2}{*} {POGP} & 0.542 & 0.164 & 0.562 & 0.186 \\
& (0.014) & (0.007) & (0.018) & (0.008) \\
\multirow{2}{*} {PMGP} & 0.564 & 0.174 & 0.574 & 0.202 \\
& (0.014) & (0.009) & (0.015) & (0.008) \\
\multirow{2}{*} {GP} & 0.551 & 0.174 & 0.586 & 0.207 \\
& (0.010) & (0.009) & (0.013) & (0.009) \\
\multirow{2}{*} {LG} & 0.568 & 0.195 & 0.584 & 0.222 \\
& (0.014) & (0.009) & (0.013) & (0.010) \\
\multirow{2}{*} {CT} & 0.669 & 0.273 & 0.661 & 0.284 \\
& (0.008) & (0.007) & (0.013) & (0.005) \\
\bottomrule
\end{tabular}
\end{center}
\end{table}

\subsection{Discussion}

In summary, our experiments show that modeling trust correlations across tasks improves predictions. Our Bayesian and neural models achieve better performance than existing approaches that treat trust as a single scalar value. To be clear, neither model attempts to represent exact trust processes in the human brain; rather, they are computational analogues. Both modeling approaches offer conceptual frameworks for capturing the principles of trust formation and transfer. From one perspective, the GP model extends the single global trust variable used in \cite{Chen2018Hri} and ~\cite{xu2015optimo} to a \emph{collection} of latent trust variables. In the following, we highlight matters relating to the learned feature projections, {changes in trust during a task}, and the neural-GP trust updates.

\begin{figure}
\centering
\includegraphics[width=0.50\columnwidth]{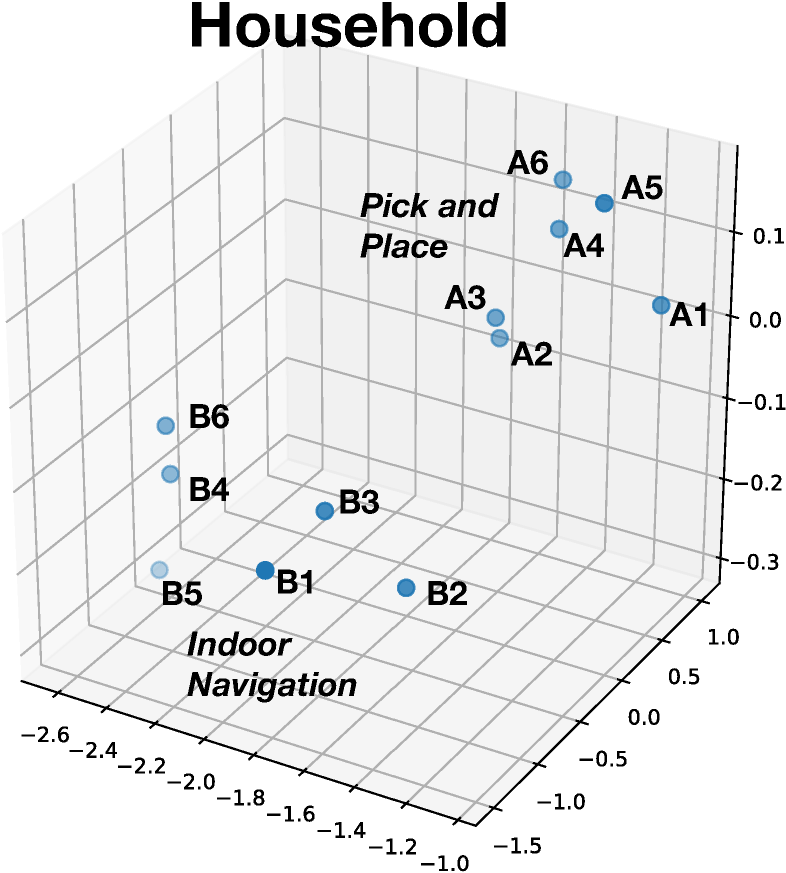}
	\caption{The task space for the Household domain where each point is a task. Tasks of a similar type and difficulty are clustered together; tasks labelled A correspond to Pick-and-Place tasks and B are Indoor Navigation tasks. The lower-numbered tasks (1-3) were considered easier by participants.}
	\label{fig:LSHousehold}
\end{figure}

\begin{figure}
\centering
\includegraphics[width=0.50\columnwidth]{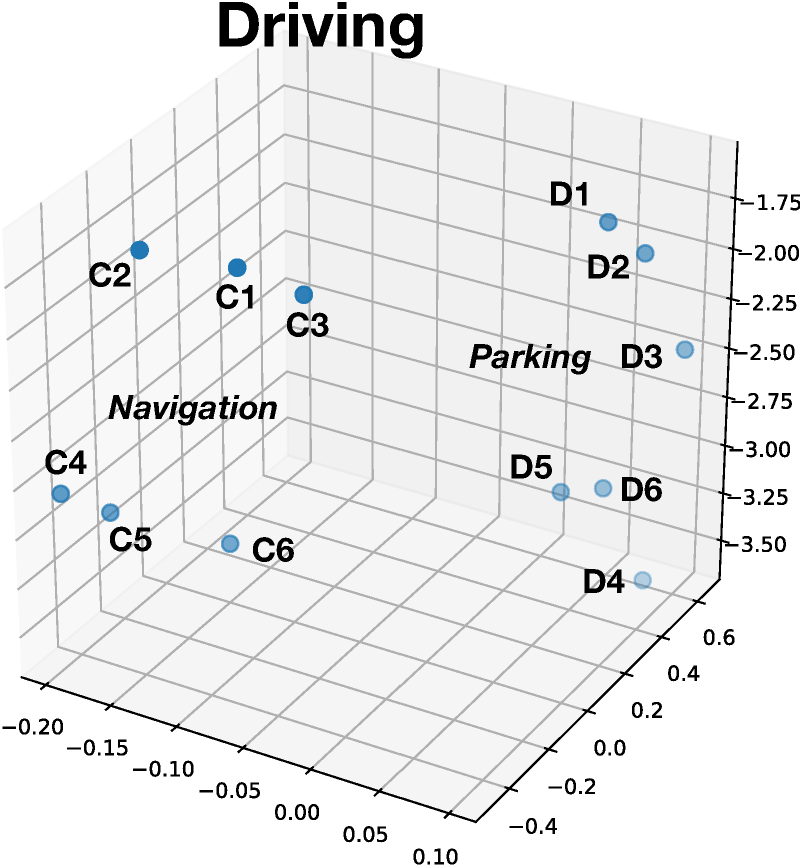}
	\caption{The task space for the Driving domain. Similar to Fig. \ref{fig:LSHousehold} above, similar tasks are closer together. Points C corresponds to Navigation tasks (e.g., lane merging) and D are Parking tasks. The lower-numbered tasks (1-3) were considered easier by participants.}
	\label{fig:LSDriving}
\end{figure}

\paragraph{Learned Trust-Task Spaces.} Although the neural and Bayesian models differ conceptually and in details, they both leverage upon the idea of a vector task space $Z \subseteq \mathbb{R}^k$ in which similarities---and thus, trust transfer---between tasks can be easily computed. For the RNN, $Z$ is a dot-product space. For the GP, similarities are computed via the kernel function; the kernel linearly projects the task features into a lower dimensional space ($\mat{z} = \boldsymbol{\Lambda}\mat{x}$) where an anisotropic squared exponential kernel is applied. As an example, Figs. \ref{fig:LSHousehold} and \ref{fig:LSDriving} show the learned GPNN latent task points for the Household and Driving domains respectively; we observe that tasks in the same group are clustered together. Furthermore, the easy and difficult tasks within each task group are also positioned closer together. This structure is consistent with the use of a squared exponential kernel where distances between the latent points determine covariance; the closer the points (more similar), the similar the latent function value at those points.

\paragraph{Generalization across Task Word Descriptions.} In our experiments, we used word vector representations as task features, which we found to enable reasonable generalization across similar descriptions. For example, after seeing the robot successfully navigate around obstacles, but \emph{failing} to pick up a lemon, the model predicts sensible trust values for the following tasks:
\begin{itemize}
	\item ``\emph{Navigate while following a path}'': 0.81
	\item ``\emph{Go to the table}'': 0.86
	\item ``\emph{Pick up a banana}'': 0.61
\end{itemize}
We posit that this results from the fact that vector-based word representations are generally effective at capturing synonyms and word relationships. {Given a  latent space with sufficiently large dimensionality, we expect the model to scale to a larger number of task categories and domains; there is evidence that  moderately-sized latent spaces ($<1000$) yield accurate  models for complex tasks such as language translation}~\citep{Sutskever2014} {and image captioning}~\citep{ren2017deep}. {Given longer task descriptions,  more sophisticated techniques from NLP} \citep[e.g.,][]{bowman2016generating} {beyond the simple averaging used in our experiments can be adapted to construct usable task features of reasonable length.}

A prevailing issue is that the current word/sentence representations may not distinguish subtle semantics, e.g., the task features lack a notion of physical constraints. As such, the model may make unreasonable predictions when given task descriptions that are syntactically similar but semantically different. As an example, the same model predicts the human highly trusts the robot's capability to ``\emph{Navigate to the moon}'' ($\tau^a = 0.83$). To remedy this issue, we can use alternative features; more informative vector-based features can be used without changing the methods described. Applying structured feature representations (e.g., graphs) would require different  kernels and embedding techniques. Future work may also examine more sophisticated hierarchical space representations.


\subsubsection{Trust Variations within a Task.} {The presented computational models are ``event-based'' in  that trust was updated after each complete task performance. However, prior work has shown that trust can change even as the task is being carried out}~\citep{Desai13,yang2017evaluating}. {To accommodate intra-task trust variability, the presented Bayesian and neural models can be easily altered to be updated at  user-defined intervals with a corresponding observation. These modifications and studies to validate such models would make for interesting future work.}

\begin{figure}
\centering
\includegraphics[width=0.6\columnwidth]{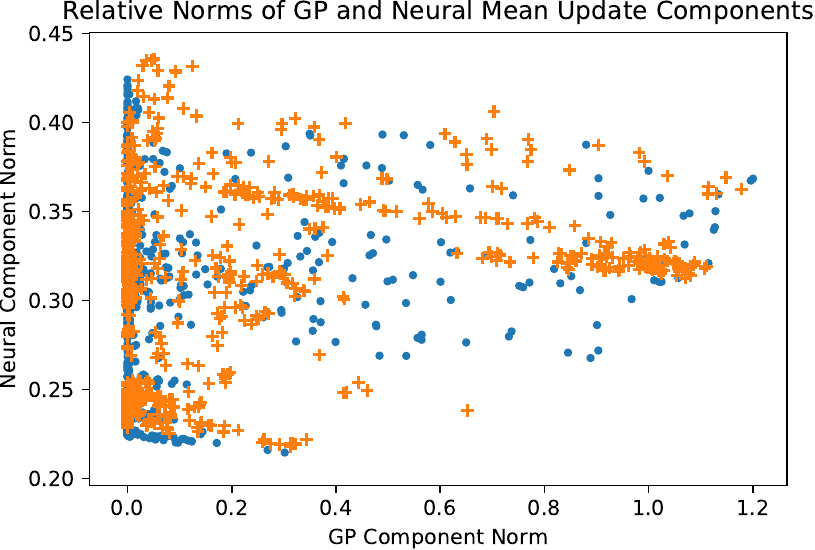}
	\caption{The relative norms of the GP component $\eta_{\textrm{GP}}$ (x-axis) and neural component $\eta_{\textrm{NN}}$ (y-axis) for updates across randomly sampled tasks. Blue points are for the Household domain and orange +'s for Driving. There is a general positive correlation between the norms, but the relationship is nonlinear.}
	\label{fig:GPNNnorms}
\end{figure}

\paragraph{GP-Neural Updates.} Finally, we sought to better understand the relationship between the Bayesian and neural components of the GPNN mean update. Was the neural network $u(\cdot)$ only making minor ``corrections'' to the trust update or was it playing a larger role? To answer this question, we compared the relative norms of the second term ($\eta_{\textrm{GP}} = {\| \gamma_1(\mathbf{C}_{t-1} \mathbf{k}_{t} + \mathbf{e}_{t})\|}/{\|\boldsymbol{\alpha}_{t-1} \|}$) and third term ($\eta_{\textrm{NN}}  = {u(\boldsymbol{\alpha}_{t-1}, \mathbf{C}_{t-1}\mathbf{k}_{t}, \boldsymbol{\Lambda}\mat{x}_{t-1}, c^\robot_{t-1})}/{\|\boldsymbol{\alpha}_{t-1} \|}$) on the RHS of Eq. (\ref{eq:bayesnnupdate}) during updates across randomly sampled tasks. Fig. \ref{fig:GPNNnorms} shows a scatter plot of the relative norms of both components. We find a positive correlation (Kendall tau $= 0.2$, $p$-value $=10^{-39}$), but the relationship is clearly nonlinear. Interestingly, $\eta_{\textrm{NN}}$ could be relatively large when the $\eta_{\textrm{GP}}$ was close to zero, indicating that neural component was playing a significant role in the trust update. We also experimented with completely removing the Bayesian portion of the update, but this modification had poorer performance, potentially due to limited data. This suggests that trust is not purely Bayesian and a non-trivial correction is needed to achieve better performance. 

\section{Conclusion}
\label{sec:conclusion}
This paper takes a first step towards conceptualizing and formalizing predictive of  human trust in robots across multiple tasks. It presents findings from a human-subjects study in two separate domains and shows the effects of task similarity and difficulty on trust formation and transfer. 

The experimental findings leads to three novel models that capture the form and evolution of trust as a latent function. Our experiments show that the function-based models achieved better predictive performance on both unseen participants and unseen tasks. These results indicate that (i) a task-dependent functional trust model more accurately reflects human trust across tasks, and (ii) it is possible to accurately predict trust transfer by leveraging upon a shared task space representation and update process. 

Formalizing trust as a function opens up several avenues for future research. In particular, we aim to  fully exploit this characterization by incorporating other contexts. Does trust transfer when the environment changes  substantially  or a new, but similar robot appears? Proper experimental design to elicit and measure trust is crucial. Our current experiments employ relatively short interactions with the robot and rely on subjective self-assessments. Future experiments could employ behavioral measures, such as operator take-overs and longer-term interactions where trust is likely to play a more significant role. 

This work limits the  investigation to trust resulting from complete observations of the robot's performance/capabilities. {However, in real-world collaborative settings, the human user may not observe all the robot's successes or failures: how humans infer the robot's performance under conditions of partial observability remains an interesting open question.} It is also essential to examine trust in the robot's ``intention'', e.g., its policy~\citep{Huang2018} and decision-making process. Arguably, trust is most crucial in new and uncertain situations whereby both robot capability and intention can influence outcomes. In very recent work~\citep{Xie2019Hri}, we have begun to examine how human mental models of  these factors influence decisions to trust robots. 

Finally, it is important to investigate how these trust models  enhance human-robot interaction. {In our current experiments, the human does not get involved in task completion. Our trust model can be used without modification in the collaborative setting where the human and the robot work together to complete a task, provided that the shared goal is unaffected by the change in trust.} Embedding trust models in a decision theoretic framework enables a robot to adapt its behavior according to a human teammate's trust and as a result, promotes fluent long-term collaboration. We have begun a preliminary investigation using trust transfer models in a Partially-observable Markov Decision Process (POMDP), extending the work in ~\citep{Chen2018Hri}. {We are particularly interested in how trust models  impacts decision-making in assistive tasks}~\citep{gombolay2018robotic,Soh2015}.

\section*{Acknowledgements}
This work was supported in part by a NUS Office of the Deputy President (Research and Technology) Startup Grant and in part by MoE AcRF Tier 2 grant MOE2016-T2-2-068. Thank you to Indu Prasad for her help with the data analysis.

\bibliographystyle{plainnat}
\small 

\bibliography{transfer_rss2018}

\end{document}